\pdfoutput=1
\documentclass[11pt]{article}

\usepackage[final]{acl}

\usepackage{times}
\usepackage{latexsym}

\usepackage[T1]{fontenc}
\usepackage[utf8]{inputenc}

\usepackage{microtype}

\usepackage{inconsolata}

\usepackage{graphicx}

\usepackage{multirow}
\usepackage{threeparttable}
\usepackage{booktabs}
\usepackage{amsmath}
\usepackage{amssymb}
\usepackage[many]{tcolorbox}
\usepackage{xcolor}
\usepackage{xspace}

\newcommand{\method}{\textsc{HintQA}\xspace}

\title{Exploring Hint Generation Approaches in Open-Domain Question Answering}

\author{Jamshid Mozafari \\
  University of Innsbruck \\
  \texttt{jamshid.mozafari@uibk.ac.at} \\\And
  Abdelrahman Abdallah\footnotemark[1] \\
  University of Innsbruck \\
  \texttt{abdelrahman.abdallah@uibk.ac.at} \\\AND
  Bhawna Piryani\thanks{\, Equal contribution.} \\
  University of Innsbruck \\
  \texttt{bhawna.piryani@uibk.ac.at} \\\And
  Adam Jatowt \\
  University of Innsbruck \\
  \texttt{adam.jatowt@uibk.ac.at} \\}

\begin{document}
\maketitle

\begin{abstract}
    Automatic Question Answering (QA) systems rely on contextual information to provide accurate answers. Commonly, contexts are prepared through either retrieval-based or generation-based methods. The former involves retrieving relevant documents from a corpus like Wikipedia, whereas the latter uses generative models such as Large Language Models (LLMs) to generate the context. In this paper, we introduce a novel context preparation approach called \method, which employs Automatic Hint Generation (HG) techniques. Unlike traditional methods, \method prompts LLMs to produce hints about potential answers for the question rather than generating relevant context. We evaluate our approach across three QA datasets including TriviaQA, Natural Questions, and Web Questions, examining how the number and order of hints impact performance. Our findings show that the \method surpasses both retrieval-based and generation-based approaches. We demonstrate that hints enhance the accuracy of answers more than retrieved and generated contexts.
\end{abstract}
\section{Introduction}\label{s:introduction}

\begin{figure}[tb]
	\includegraphics[width=\columnwidth]{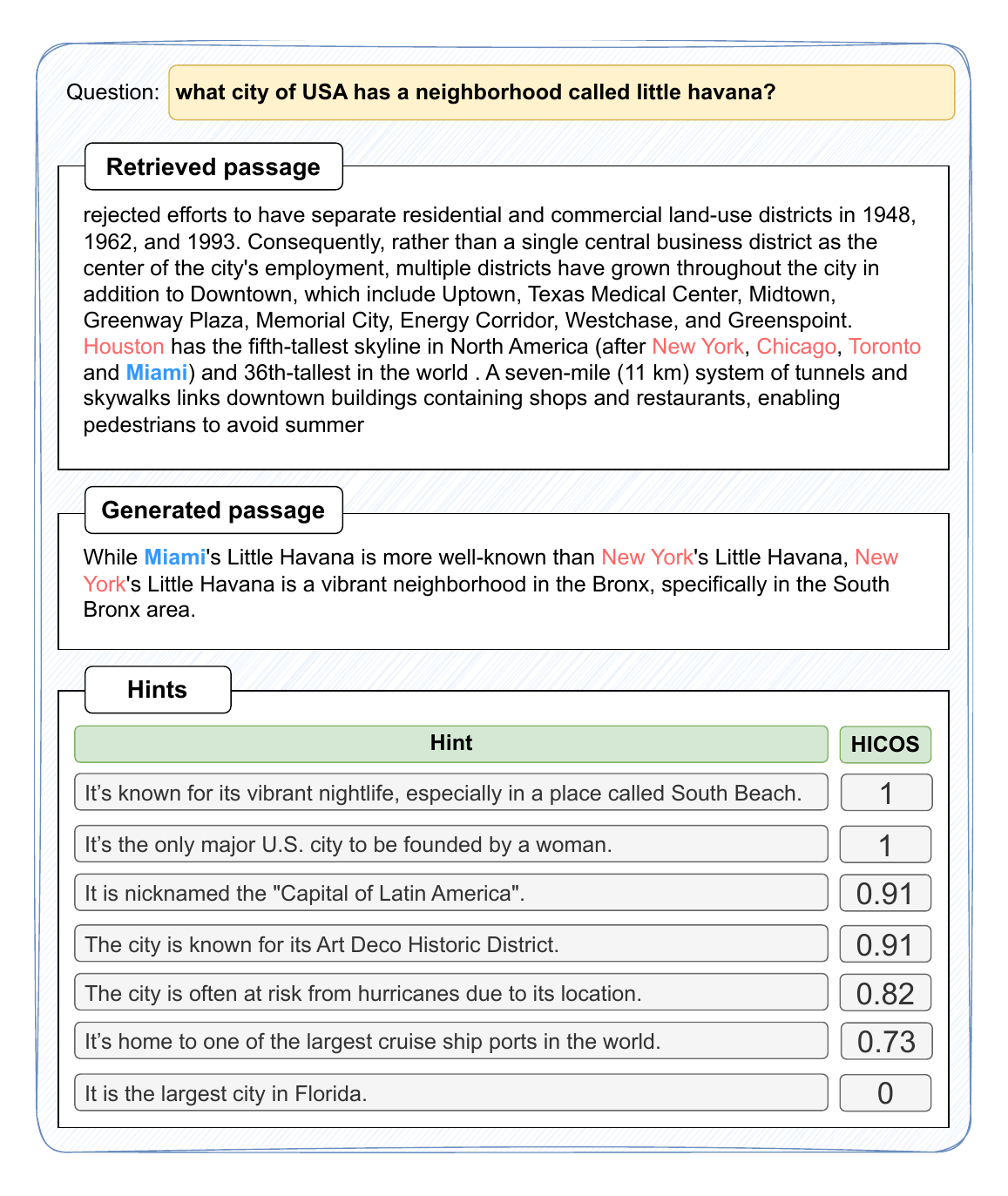}
	\caption{Example of generated hints, context produced by LLaMA-70, and a passage retrieved by MSS-DPR for a TriviaQA sample question, with convergence score (HICOS) ranging from 0 (lowest) to 1 (highest). Words in \textcolor{blue}{blue} indicate the correct answer, while those in \textcolor{red}{red} represent other potential answers.}
	\label{fig:hint_vs_generated_retrieved_passage}
\end{figure}

Automatic Question Answering (QA) systems~\citep{Abdel-Nabi2023} have recently garnered significant attention. They allow users to receive direct responses to posed questions. QA systems typically comprise three main components: Context-Preparator, Reranker, and Reader~\citep{10.1145/3560260}. The Context-Preparator component aims to supply relevant context to the user question. The Reranker then prioritizes the documents based on their relevance to the question or to potential answers~\citep{mao-etal-2021-reader}. Lastly, the Reader extracts the answer from the provided context. The Context-Preparator component is the initial step and a crucial element in QA systems. If this component fails to prepare the most relevant contexts, the entire QA system can be led astray. Therefore, the accuracy and performance of the Context-Preparator component are crucial for the overall success of QA systems.
The Context-Preparator component may belong to two primary categories: Retrieval-based and Generation-based approaches~\citep{li2024matching}.

Retrieval-based methods retrieve relevant passages from document collections, such as Wikipedia, using techniques like keyword matching~\citep{10.1007/11554028_10} or vector space models~\citep{10.1145/3196826}. A  limitation of these methods is that retrieved passages tend to be lengthy, often exceeding 100 words~\citep{karpukhin-etal-2020-dense}. Consequently, some sentences within these passages may be irrelevant to the question~\citep{mitra2017neural}. Figure~\ref{fig:hint_vs_generated_retrieved_passage} illustrates a retrieved passage where only one sentence contains the potential answers including also the correct one, while the other sentences are irrelevant.

In contrast, generation-based methods use generative models, such as large language models (LLMs)~\citep{workshop2022bloom} and Seq-to-Seq techniques~\citep{NIPS2014_a14ac55a}, to produce relevant context. A major limitation of these methods is that they typically produce only a small number of sentences as context, usually just one or two. 
When the number of sentences is small, there is a risk that the QA system could be mislead if the answer is incorrect, due to insufficient context to substantiate the answer.
Figure~\ref{fig:hint_vs_generated_retrieved_passage} also displays a generated passage consisting of only two sentences, which could mislead the Reader. This is because the correct answer appears less frequently than incorrect ones, and the scant context does not provide sufficient information for the Reader component to identify the correct answer accurately.

Our research aims to overcome the shortcomings of both retrieval-based and generation-based methods. It eliminates irrelevant sentences and provides only those containing useful information about the answer, thereby addressing a key limitation of the retrieval-based method. Additionally, we aim to expand the number of informative sentences beyond just one or two as usually is in the case of generated context, tackling a major drawback of the generation-based approach.

We present \method\footnote{The code, dataset, and experimental results are freely available at \url{https://github.com/DataScienceUIBK/HintQA}}, a novel approach that utilizes Automatic Hint Generation (HG) systems~\citep{jangra2024navigating} to generate hints as the context. This method produces multiple hints for each question and substitutes the retrieved passages and generated contexts with the generated hints.
Figure~\ref{fig:hint_vs_generated_retrieved_passage} illustrates seven generated hints, each accompanied by its computed convergence score (HICOS). The convergence score is a measure that indicates how effectively a hint can narrow down or eliminate potential answers to a given question~\citep{mozafari2024triviahg}. The hints can be then subsequently reranked based on criteria such as the aforementioned convergence score or semantic relevance, setting the stage for the Reader to discern the correct answer from the prioritized hints. To assess the effectiveness of our approach, we generate hints for each question belonging to the test sets of the TriviaQA~\citep{joshi-etal-2017-triviaqa}, Natural Questions (NQ)~\citep{kwiatkowski-etal-2019-natural}, and Web Questions (WebQ)~\citep{berant-etal-2013-semantic} datasets. Table~\ref{tbl:dataset_statistics} and Table~\ref{tbl:dataset_distribution} in Appendix~\ref{apx:dataset} indicate the statistics and distributions of these datasets. Our extensive experiments demonstrate that using hints leads to better performance than relying on retrieved passages or generated context. To sum up, we make the following contributions in this work:
\begin{itemize}
    \item We propose a novel approach for the Context-Preparator component in QA systems that is based on using hint generation techniques. 
    \item We generate and release hints along with their corresponding convergence scores 
    for the questions of the test sets of the TriviaQA, NQ, and WebQ datasets.
    \item We conduct extensive experiments on these datasets using zero and few-shot strategies across various numbers of hints and reranking methods. 
\end{itemize}
\section{Related Work}\label{s:related_work}

\subsection{Retrieval-based Methods}\label{ss:retrieval-based}
Retrieval-based methods can be divided into two primary categories: (1) Sparse retrieval and (2) Dense retrieval. Sparse retrieval methods rely on word-level matching to establish connections between vocabulary and documents. 
Notable examples are Boolean Retrieval~\citep{salton1983extended}, BM25~\citep{robertson2009probabilistic}, SPLADE~\citep{formal2021splade}, and UniCOIL~\citep{lin2021few}.
On the other hand, dense retrieval methods capture deep semantic information from documents to understand underlying semantics and improve retrieval accuracy. 
Some key examples are DPR~\citep{karpukhin-etal-2020-dense}, ANCE~\citep{xiong2020approximate}, E5~\citep{wang2022text}, and SimLM~\citep{wang-etal-2023-simlm}. 

\begin{figure*}[tb]
    \centering
	\includegraphics[width=0.8\textwidth]{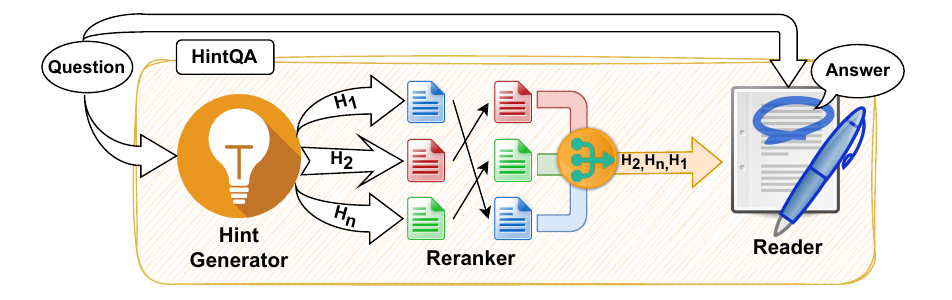}
	\caption{The \method approach, where $H_i$ denotes the $i$th hint. Initially, the Hint Generation component produces hints for the given question. These hints are then reranked and concatenated to form a context, which is subsequently passed to the Reader component to identify the answer of the question.}
	\label{fig:framwork}
\end{figure*}

\subsection{Generation-based Methods}\label{ss:generation-based}
Generation-based systems can be broadly classified into two main categories: (1) Generative document retrieval and (2) Reliable response generation. Generative document retrieval utilizes the parametric memory of generative models to retrieve relevant documents. 
Unlike retrieval-based systems, this approach depends on pre-trained generative models, such as BART~\citep{lewis-etal-2020-bart}, to produce document identifiers directly related to the question.
Some notable examples are DSI~\citep{tay2022transformer}, DynamicRetriever~\citep{zhou2023dynamicretriever}, SEAL~\citep{bevilacqua2022autoregressive}, and NCI~\citep{wang2022neural}.
Conversely, Reliable response generation methods provide a more dynamic form of information access by directly producing detailed, user-centric responses. 
Notable instances are LLaMA~\citep{brown2020language}, InstructGPT~\citep{ouyang2022training}, T5~\citep{10.5555/3455716.3455856}, PaLM~\citep{chowdhery2023palm} and Copilot\footnote{\url{https://copilot.microsoft.com/}}.

\subsection{Hint Generation}\label{ss:hint-generation}
HG systems can be categorized into two main categories: (1) Hint generation for Programming (AHGP) and (2) Hint generation for Questions (AHGQ). AHGP aims to create helpful hints for programming exercises~\citep{10.1145/2960310.2960333}. 
Some notable examples are ITAP~\citep{10.1007/978-3-642-30950-2_40} and Catnip~\citep{10.1145/3430665.3456344} systems.
In contrast, methods for AHGQ focus on generating hints for user questions rather than programming exercises~\citep{mozafari2024triviahg, jangra2024navigating}. 
\citet{10.1145/3578337.3605119} explore the use of Wikipedia for generating hints without utilizing LLMs, primarily to introduce this as a new area of research. \citet{mozafari2024triviahg} advance the field by releasing the first dedicated dataset named TriviaHG, along with a novel automatic evaluation method for assessing the quality of hints.

To the best of our knowledge, no study has yet explored the use of AGHQ approaches as the Context-Preparator component for QA systems.

\section{Method} \label{s:method}
In this section, we first explore the theoretical foundations underpinning our approach, followed by a detailed explanation of its implementation.

\begin{figure*}[tb]
	\includegraphics[width=\textwidth]{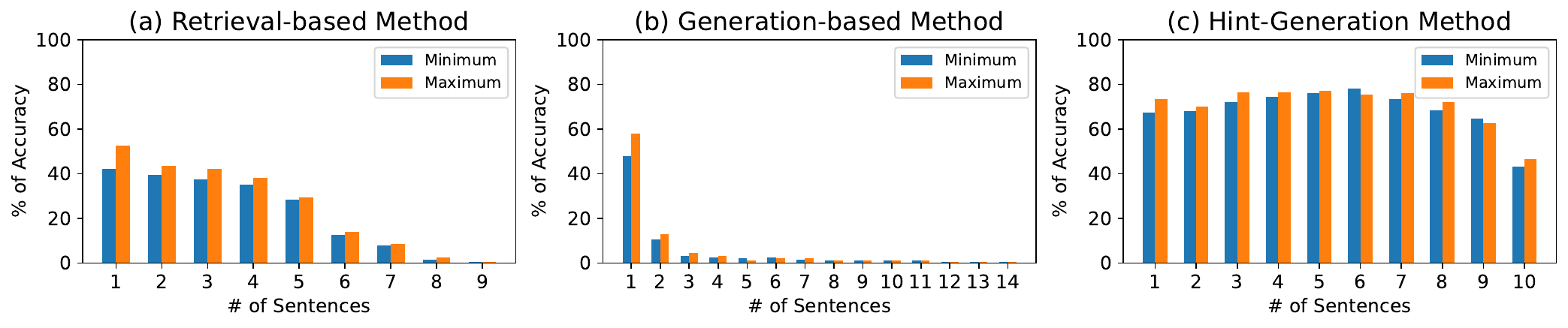}
 \caption{Accuracy results for 200 random questions from TriviaQA, NQ, and WebQ when using LLaMA-7b as the Reader and varying the numbers of context sentences. The context sentences are obtained by (a) Retrieval-based (DPR), (b) Generation-based (LLaMA-70b), and (c) Hint-Generation (HiGen-FT) methods. The \textcolor{blue}{blue} (\textcolor{red}{red}) columns indicate the accuracy when the total number of potential entities across sentences is at its minimum (maximum). The number of potential entities per sentence is calculated using HICOS approach \citep{mozafari2024triviahg}.}
	\label{fig:theorical}
\end{figure*}

\subsection{Hypothesis} \label{ss:hypothesis}

Let $q$ be a question linked to a set of candidate answers $\mathcal{A} = \{a_1, a_2, \dots, a_n\}$, such that $q \rightarrow \mathcal{A}$, which indicates that $\mathcal{A}$ is assumed to encompass all possible answers to $q$. Additionally, let $\mathcal{S} = \{s_1, s_2, \dots, s_j\}$ be the context, consisting of a series of sentences $s_i$ provided to determine the answer to $q$. Each sentence $s_i$ typically discusses or relates to certain entities or subjects, which we refer to as $\mathcal{C}'_i$. For instance, the sentence \textit{"He was a professional."} might pertain to different possible professions such as actor, painter, athlete, etc. Consequently, the set $\mathcal{C}'_i$ could encompass, in this example, various individuals from diverse occupations. However, if the question $q$ specifically inquires about just one particular profession, it is superfluous to consider all potential entities that the sentence might include. Therefore, we define $\mathcal{C}_i = \mathcal{C}'_i \cap \mathcal{A}$ to select only those entities that represent the intersection between the candidate answers for $q$ and the possible entities from $s_i$. This process assists in eliminating irrelevant entities, retaining only valid candidate answers to $q$. 

We define a score $\tau_\mathcal{S}(a)$ for a candidate answer $a$ within the context $\mathcal{S}$ to represent how well $a$ scores as a candidate answer in the context $\mathcal{S}$. It counts the number of supporting sentences for the candidate answer $a$ among all sentences in $\mathcal{S}$:
\begin{equation}
    \label{eq:score}
    \tau_\mathcal{S}(a) = \frac{\sum_{s \in \mathcal{S}} \chi_{\mathcal{C}_s}(a)}{|\mathcal{S}|}
\end{equation}
where $|\mathcal{S}|$ indicates the number of sentences within $\mathcal{S}$, and $\mathcal{C}_s$ identifies the valid candidate answer set associated with sentence $s$. The function $\chi_{\mathcal{C}_s}(a)$ is to determine whether a candidate answer $a$ is a member of the candidate answer set $\mathcal{C}_s$:
\begin{equation} 
    \label{eq:inclusion}
    \chi_{\mathcal{C}_s}(a) = 
    \begin{cases} 
    1 & \text{if } a \in \mathcal{C}_s \\
    0 & \text{if } a \notin \mathcal{C}_s \\
    \end{cases}
\end{equation}

The candidate answer $a$ with the highest $\tau_\mathcal{S}(a)$ across the context $\mathcal{S}$ is proposed as the most likely correct answer:

\begin{equation}
    \label{eq:answer}
    a^* = \arg\max_{a \in \mathcal{A}} \tau_\mathcal{S}(a)
\end{equation}

Let's consider an example as follows. Suppose the question $q$ is: \textit{"What city in the USA has a neighborhood called Little Havana?"}. 
And suppose the context $\mathcal{S}$ %could for instance 
consists of two sentences $s_1$ (\textcolor{red}{red}) and $s_2$ (\textcolor{blue}{blue}):

\begin{tcolorbox}[fontupper=\ttfamily\small]
\textcolor{red}{The city is often at risk from hurricanes due to its location.} \textcolor{blue}{Additionally, it’s the only major U.S. city to be founded by a woman.}
\end{tcolorbox}

The entities 
supported by $s_1$ are $\mathcal{C}'_1 = \{\textit{San Juan, Kingston, Miami, New York, \dots}\}$, and ones by $s_2$ are $\mathcal{C}'_2 = \{\textit{Miami}\}$. 
Let us also suppose that the following candidate answers are possible for q: $\mathcal{A} = \{\textit{Houston, Miami, New York}\}$. 
Thus, the intersecting sets are $\mathcal{C}_1 = \mathcal{C}'_1 \cap \mathcal{A} = \{\textit{Miami, New York}\}$ and $\mathcal{C}_2 = \mathcal{C}'_2 \cap \mathcal{A} = \{\textit{Miami}\}$. We calculate the score $\tau_\mathcal{S}$ for \textit{Miami} using Eq.~\ref{eq:score}:

%\noindent
{\small
\begin{align}
    \label{eq:score_for_miami}
    \tau_\mathcal{S}(\text{Miami}) = \frac{\chi_{\mathcal{C}_1}(\text{Miami}) + \chi_{\mathcal{C}_2}(\text{Miami})}{|\mathcal{S}|} = \frac{2}{2} = 1
\end{align}
}
The scores for \textit{Houston} and \textit{New York} are $0$ and $0.5$, respectively. Thus, according to Eq.~\ref{eq:answer}, the most likely correct answer to $q$ given the context is \textbf{Miami} as supported by most of the sentences.

We believe that a context supporting more potential entities in its sentences can improve the performance of QA systems. As shown in Figure~\ref{fig:theorical}, the \textit{Maximum} column illustrates that when the total number of potential entities across sentences is highest, the accuracy exceeds that observed with the lowest count. Figure~\ref{fig:theorical}b also demonstrates how a scarcity of potential entities can mislead the QA system. As discussed in Section~\ref{s:introduction}, this issue is especially common in generation-based methods, which frequently produce contexts with a small number of sentences.

Moreover, Figure~\ref{fig:theorical}a shows that additional sentences can impair QA system performance if the sentences are irrelevant.
The figure demonstrates a correlation between an increase in irrelevant sentences and a decrease in accuracy. This presents a frequent challenge for retrieval-based methods, which are prone to including irrelevant sentences in the passages they retrieve.

Nevertheless, Figure~\ref{fig:theorical}c demonstrates that the results of the HG method 
can effectively guide the QA system toward the correct answer. Table~\ref{tbl:case_study_candidates} in Appendix~\ref{apx:case_study} provides also some generated hints and their supported candidate answers.

\subsection{Implementation} \label{ss:implementation}
To implement our approach, we adapt the method introduced by~\citet{mozafari2024triviahg} for generating ten hints, modifying their original prompt. While they implemented an answer-aware approach, we take an answer-agnostic approach since the correct answer is unknown. Note that the HG method occasionally generates hints where the answer is replaced with blanks to prevent answer leakage. We remove such hints, which could result in some questions having fewer than 10 hints. The prompt we use to generate hints is as follows:

\begin{tcolorbox}[fontupper=\ttfamily\small]
Generate 10 concise and relevant hint sentences for the following question. List the hints without revealing the answers within them.
\end{tcolorbox}

After generating hints, we rerank the created hints based on the HICOS score and concatenate them to make a new context. Finally, we pass the generated context to a Reader to answer a given question. Figure~\ref{fig:framwork} shows \method approach. We utilize the following prompt in the Reader to extract the answer from the context:

\begin{tcolorbox}[fontupper=\ttfamily\small]
According to the following context, answer the question: \\
Context: \textbf{Provided Context} \\
Question: \textbf{Given Question} \\
Answer: \textbf{Here is the answer}
\end{tcolorbox}

\section{Experimental Setup}\label{s:experiments}

\subsection{Datasets}\label{ss:dataset}
Our evaluation is conducted using three diverse datasets: TriviaQA~\citep{joshi-etal-2017-triviaqa}, NQ (Natural Questions)~\citep{kwiatkowski-etal-2019-natural}, and WebQ~\citep{berant-etal-2013-semantic}.
TriviaQA dataset comprises a comprehensive collection of trivia questions, which have been curated from various trivia and quiz-league websites.
NQ has been constructed from Google Search queries, providing a realistic set of questions people ask. The answers to these questions are drawn as specific spans or segments from Wikipedia articles.
WebQ dataset consists of questions sourced from the Google Suggest API, which generates predictive search suggestions based on user input. The answers are tied to entities within Freebase~\citep{10.1145/1376616.1376746}.
A more detailed description of dataset statistics, their splits (Table~\ref{tbl:dataset_statistics}), and distributions based on the question type (Table~\ref{tbl:dataset_distribution}) can be found in Appendix~\ref{apx:dataset}.

\begin{table}[t]
    \small
	\centering
	\begin{tabular}{@{}l|lll@{}}
		\toprule
		Method     & TriviaQA & NQ     & WebQ   \\ \midrule
		BM25       & 117.15   & 114.93 & 114.24 \\
		DPR        & 118.66   & 110.97 & 114.56 \\
		Contriever & 117.41   & 107.47 & 113.69 \\
		MSS        & 118.62   & 113.44 & 117.25 \\
		MSS-DPR    & 118.35   & 109.56 & 115.66 \\ \midrule
		LLaMA-70b  & 50.34    & 61.52  & 75.93  \\ \midrule
		HiGen-FT   & 73.54    & 96.13  & 90.43  \\
		HiGen-Va   & 96.85    & 106.78 & 93.02  \\ \bottomrule
	\end{tabular}
	\caption{Comparison of average lengths of hints, generated contexts, and retrieved passages based on the number of words.}
	\label{tbl:avg_length_hint_gen_ret}
\end{table}
\begin{table}[htb]
	\resizebox{\columnwidth}{!}{%
		\begin{threeparttable}
			[b]
			\begin{tabular}{@{}lllllll@{}}
				\toprule \multicolumn{1}{c|}{\multirow{2}{*}{Method}} & \multicolumn{2}{c|}{TriviaQA\tnote{1}} & \multicolumn{2}{c|}{NQ\tnote{2}}    & \multicolumn{2}{c}{WebQ\tnote{3}} \\
				\multicolumn{1}{c|}{}                         & EM             & \multicolumn{1}{l|}{F1}             & EM             & \multicolumn{1}{l|}{F1}             & EM             & F1             \\
				\midrule \multicolumn{7}{c}{\textit{Zero-Shot}}           \\
				\midrule \multicolumn{1}{l|}{BM25}                     & 23.28          & \multicolumn{1}{l|}{27.22}          & 3.55           & \multicolumn{1}{l|}{5.62}           & 10.97          & 18.54          \\
				\multicolumn{1}{l|}{Contriever}               & 18.13          & \multicolumn{1}{l|}{22.29}          & 1.94           & \multicolumn{1}{l|}{3.66}           & 8.17           & 14.05          \\
				\multicolumn{1}{l|}{DPR}                      & 23.22          & \multicolumn{1}{l|}{27.7}           & 2.3            & \multicolumn{1}{l|}{3.93}           & 11.71          & 19.43          \\
				\multicolumn{1}{l|}{MSS}                      & 18.15          & \multicolumn{1}{l|}{22.35}          & 1.97           & \multicolumn{1}{l|}{3.58}           & 9.94           & 17.24          \\
				\multicolumn{1}{l|}{MSS-DPR}                  & 18.14          & \multicolumn{1}{l|}{22.23}          & 4.24           & \multicolumn{1}{l|}{6.53}           & 11.17          & 18.71          \\
				\midrule \multicolumn{1}{l|}{LLaMA-70b} & 21.45          & \multicolumn{1}{l|}{26}             & 3.88           & \multicolumn{1}{l|}{6.23}           & 12.11          & 20.27          \\
                \midrule \multicolumn{1}{l|}{HiGen-Va}                  & 22.01          & \multicolumn{1}{l|}{26.5}           & 9.06           & \multicolumn{1}{l|}{12.54}          & 13.88          & 21.74          \\
				\multicolumn{1}{l|}{HiGen-FT}                 & \textbf{23.55} & \multicolumn{1}{l|}{\textbf{28.03}} & \textbf{10.89} & \multicolumn{1}{l|}{\textbf{14.85}} & \textbf{14.96} & \textbf{23.08} \\
				\midrule \multicolumn{7}{c}{\textit{Few-Shot}}            \\
				\midrule \multicolumn{1}{l|}{BM25}                     & 25.78          & \multicolumn{1}{l|}{30.29}          & 4.6            & \multicolumn{1}{l|}{7.33}           & 11.17          & 18.93          \\
				\multicolumn{1}{l|}{Contriever}               & 21.48          & \multicolumn{1}{l|}{25.87}          & 2.47           & \multicolumn{1}{l|}{4.21}           & 7.53           & 13.49          \\
				\multicolumn{1}{l|}{DPR}                      & 25.02          & \multicolumn{1}{l|}{29.49}          & 3.24           & \multicolumn{1}{l|}{5.09}           & 11.37          & 19.37          \\
				\multicolumn{1}{l|}{MSS}                      & 20.89          & \multicolumn{1}{l|}{25.27}          & 2.85           & \multicolumn{1}{l|}{4.75}           & 10.33          & 17.99          \\
				\multicolumn{1}{l|}{MSS-DPR}                  & 20.92          & \multicolumn{1}{l|}{25.19}          & 4.79           & \multicolumn{1}{l|}{7.69}           & 11.47          & 19.81          \\
				\midrule \multicolumn{1}{l|}{LLaMA-70b} & 23.64          & \multicolumn{1}{l|}{28.86}          & 5.1            & \multicolumn{1}{l|}{7.9}            & 9.4            & 17.86          \\
                \midrule \multicolumn{1}{l|}{HiGen-Va}                  & 34.19          & \multicolumn{1}{l|}{39.74}          & 12.85          & \multicolumn{1}{l|}{18.06}          & 18.9           & 28.97          \\
				\multicolumn{1}{l|}{HiGen-FT}                 & \textbf{38.54} & \multicolumn{1}{l|}{\textbf{44.29}} & \textbf{16.68} & \multicolumn{1}{l|}{\textbf{22.64}} & \textbf{24.11} & \textbf{34.52} \\
				\bottomrule
			\end{tabular}
			\begin{tablenotes}
				\item[1] Zero-Shot$\rightarrow$ HiGen-Va: 10 Def, HiGen-FT: 10 Def \\ Few-Shot$\rightarrow$
				HiGen-Va: 5 Conv, HiGen-FT: 7 Def 
                \item[2] Zero-Shot$\rightarrow$ HiGen-Va: 10
				Def, HiGen-FT: 10 Def \\ Few-Shot$\rightarrow$ HiGen-Va: 5 Conv, HiGen-FT:
				7 Def 
                \item[3] Zero-Shot$\rightarrow$ HiGen-Va: 2 Conv, HiGen-FT: 10 Def \\
				Few-Shot$\rightarrow$ HiGen-Va: 5 Conv, HiGen-FT: 7 Conv
			\end{tablenotes}
		\end{threeparttable}%
	}
    \caption{The results for \textbf{T5-3b} used as the reader, utilizing zero-shot and few-shot strategies. The footnotes provide information on the optimal number of hints and the ranking method chosen to achieve the best results for each learning strategy and hint generation method.}
    \label{tbl:result_t5}
\end{table}

\subsection{Baseline Models}\label{ss:baselines}
To compare our approach with other methods, we choose several retrieval-based and generative-based methods as baselines.

\textbf{BM25}~\citep{robertson2009probabilistic} is a probabilistic retrieval model that employs term frequency (TF) and inverse document frequency (IDF) metrics to assess the relevance of documents based on the common words in the question and the documents.
\textbf{Contriever}~\citep{izacard2022unsupervised} is an unsupervised framework designed for pre-training models for retrieval tasks, utilizing contrastive learning techniques.
\textbf{MSS}~\citep{sachan-etal-2021-end} is a dense retrieval model trained to predict masked salient spans, such as named entities, using a reader network.
\textbf{DPR}~\citep{karpukhin-etal-2020-dense} uses annotated question-context paragraphs and hard negative examples to train a supervised dense retriever.
\textbf{MSS-DPR}~\citep{sachan-etal-2021-end} enhances the performance of DPR by initially pre-training the dense retriever with MSS. This is followed by supervised fine-tuning in the style of DPR. \textbf{LLaMA-v2}~\citep{touvron2023llama} is an advanced LLM tailored for scalable natural language processing tasks, providing exceptional efficiency in generating context. 

We employ the preprocessed English Wikipedia dump, provided by~\citet{karpukhin-etal-2020-dense}, as a source for our evidence passages in retrieval-based methods. We also utilize the first top retrieved passage for the Reader. We use the LLaMA-70b as the generation-based baseline because we use LLaMA-70b in our implementation~(Section \ref{ss:implementation}) as the core of an HG system. Therefore, it is reasonable to compare the \method method directly with LLaMA-70b to ensure a fair assessment.

\subsection{Hint Generation Methods}\label{ss:hint_generation_methods}
We employ two versions of HG systems to create hints for questions: The vanilla version (HiGen-Va) and the finetuned version (HiGen-FT). In the HiGen-Va, the LLaMA-70b model is simply prompted to generate hints for a specific question. For the HiGen-FT, we first finetune the LLaMA-70b model using the TriviaHG dataset~\citep{mozafari2024triviahg}, and then prompt it to generate hints. For the detailed statistics of the TriviaHG dataset, readers are referred to Table~\ref{tbl:triviahg_statistics} in Appendix~\ref{apx:dataset}.

Additionally, we explore three different reranking methods for reranking hints: Default (Def), RankT5 (T5), and Convergence (Conv). The Default order refers to the sequence in which the hints are originally generated by the HG system. The RankT5 method rearranges hints through pairwise and listwise ranking techniques employing the T5 model~\citep{10.1145/3539618.3592047}. Lastly, the Convergence method sorts the hints according to the HICOS score in descending order.

We also investigate the impact of using various quantities of hints to prepare context. In our experiments, we concatenate the first 2, 5, 7, or 10 hints in various sequences to generate a comprehensive context for the Reader component. 
This approach allows us to assess how the number and order of hints influence the effectiveness and performance of the QA system. To compare results, we use the metrics mentioned in Appendix~\ref{apx:metrics}.

\subsection{Readers}\label{ss:readers}
We utilize two distinct language models, T5-3b~\citep{10.5555/3455716.3455856} and LLaMA-7b~\citep{touvron2023llama}, as the Reader component in our system. In addition to employing these models, we incorporate techniques such as Zero-Shot and Few-Shot\footnote{The choice to limit the number of shots to only 5 in few-shot learning is motivated by the high cost associated with exploring various shot values.} to enhance their capability to handle tasks with limited direct training on specific tasks. This setup allows us to explore the effectiveness of these models in adapting to new data and challenges using minimal examples.

\begin{table}[tb]
	\resizebox{\columnwidth}{!}{%
		\begin{threeparttable}
			[b]
			\begin{tabular}{@{}l|lllllll@{}}
				\toprule Method       & ACC            & EM            & F1            & PR            & RC             & CON            & BERT           \\
				\midrule \multicolumn{8}{c}{\textit{Zero-Shot}} \\
				\midrule BM25                     & 34.21          & 0             & 7.67          & 4.56          & 36.2           & 38.98          & 69.29          \\
				Contriever               & 20.64          & 0             & 5.57          & 3.28          & 30.71          & 26.47          & 67.13          \\
				DPR                      & 31.19          & 0             & 7.5           & 4.47          & 35.12          & 37.03          & 69.22          \\
				MSS                      & 20.38          & 0             & 5.43          & 3.19          & 30.4           & 26.11          & 67.06          \\
				MSS-DPR                  & 19.73          & 0             & 5.58          & 3.27          & 30.67          & 26.43          & 67.2           \\
				\midrule LLaMA-70b     & 47.3           & 0             & \textbf{9.11} & \textbf{5.44} & \textbf{42.57}          & 55.32          & 70.77          \\
                \midrule HiGen-Va\tnote{1}         & \textbf{59.06} & 0             & 8.04          & 4.75          & 41.51          & 54.74          & 70.35          \\
				HiGen-FT\tnote{2}        & 54.97          & 0             & 8.96          & 5.33          & 42.21 & \textbf{60.93} & \textbf{71.4} \\
				\midrule \multicolumn{8}{c}{\textit{Few-Shot}}  \\
				\midrule BM25                     & 40.5           & 38.15         & 46.7          & 46.2          & 52.8           & 51.06          & 83.32          \\
				Contriever               & 31.62          & 33.54         & 40.4          & 39.9          & 47.31          & 42.86          & 80.46          \\
				DPR                      & 36.29          & 37.15         & 45.3          & 44.8          & 51.06          & 49.16          & 82.91          \\
				MSS                      & 31.56          & 33.99         & 41.1          & 40.7          & 47.84          & 43.41          & 80.66          \\
				MSS-DPR                  & 31.96          & 32.69         & 39.9          & 39.4          & 46.95          & 42.43          & 80.2           \\
                \midrule LLaMA-70b & 52.59          & 41.26         & 48.7          & 48.6          & 52.59          & 51.58          & 83.3           \\
				\midrule HiGen-Va\tnote{3}         & 57.71          & 50.76         & 60.6          & 60.4          & 65.12          & 65.92          & 88.61          \\
				HiGen-FT\tnote{1}        & \textbf{58.06} & \textbf{54.6} & \textbf{64.7} & \textbf{64.8} & \textbf{69.53} & \textbf{70.15} & \textbf{89.89} \\
				\bottomrule
			\end{tabular}
			\begin{tablenotes}
				\item[1] 7 hints, Convergence reranking. 
                \item[2] 10 hints, Default reranking. 
                \item[3] 5 hints, Convergence
				reranking.
			\end{tablenotes}
		\end{threeparttable}%
	}
    \caption{The results for \textbf{LLaMA-7b} used as the reader on \textbf{TriviaQA}, using zero-shot and few-shot strategies.The footnotes provide information on the optimal number of hints and the ranking method chosen to achieve the best results for each learning strategy and hint generation method.}
    \label{tbl:result_llama_trivia}
\end{table}
\begin{table}[htb]
	\resizebox{\columnwidth}{!}{%
		\begin{threeparttable}
			[b]
			\begin{tabular}{@{}l|lllllll@{}}
				\toprule Method       & ACC            & EM             & F1            & PR            & RC             & CON            & BERT          \\
				\midrule \multicolumn{8}{c}{\textit{Zero-Shot}} \\
				\midrule BM25                     & 23.38          & 0              & 2.72          & 1.54          & 19             & 15.21          & 63.14         \\
				Contriever               & 11.52          & 0              & 1.84          & 1.03          & 15.71          & 10             & 61.14         \\
				DPR                      & 11.36          & 0              & 1.77          & 0.99          & 15.55          & 9.78           & 61.04         \\
				MSS                      & 11.44          & 0              & 1.67          & 0.94          & 14.75          & 9.36           & 60.94         \\
				MSS-DPR                  & 23.21          & 0              & 2.94          & 1.66          & 21.16          & 17.73          & 63.96         \\
				\midrule LLaMA-70b & 37.73          & 0              & 3.88          & 2.2           & \textbf{31.98} & 31.97          & 65.31         \\
                \midrule HiGen-Va\tnote{1}         & \textbf{51.11} & 0              & 3.44          & 1.95          & 25.71          & 26.2           & 64.97         \\
				HiGen-FT\tnote{1}        & 49.26          & 0              & \textbf{4.38} & \textbf{2.5}  & 26.96          & \textbf{33.19} & \textbf{66.8} \\
				\midrule \multicolumn{8}{c}{\textit{Few-Shot}}  \\
				\midrule BM25                     & 36.65          & 10.33          & 16.6          & 16.1          & 23.28          & 19.14          & 70.32         \\
				Contriever               & 31.66          & 6.84           & 10.7          & 10.2          & 16.17          & 11.19          & 66.61         \\
				DPR                      & 31.3           & 7.15           & 11.1          & 10.6          & 16.92          & 11.63          & 66.83         \\
				MSS                      & 29.25          & 7.15           & 11.1          & 10.5          & 17.2           & 12.05          & 66.78         \\
				MSS-DPR                  & 34.35          & 10.44          & 16.4          & 15.9          & 22.81          & 18.67          & 70.24         \\
				\midrule LLaMA-70b & 50.21          & 10.55          & 16.1          & 15.9          & 21.06          & 18.34          & 68.9          \\
                \midrule HiGen-Va\tnote{2}         & 59.36          & 18.48          & 26.6          & 26.4          & 34.58          & 33.24          & 75.58         \\
				HiGen-FT\tnote{2}        & \textbf{64.43} & \textbf{20.72} & \textbf{29.5} & \textbf{29.55} & \textbf{37.19} & \textbf{36.81} & \textbf{76.7} \\
				\bottomrule
			\end{tabular}
			\begin{tablenotes}
				\item[1] 10 hints, Convergence reranking. \item[2] 7 hints, Convergence
				reranking.
			\end{tablenotes}
		\end{threeparttable}%
	}
    \caption{The results for \textbf{LLaMA-7b} used as the reader on \textbf{NQ}, utilizing zero-shot and few-shot strategies. The footnotes provide information on the optimal number of hints and the ranking method chosen to achieve the best results for each learning strategy and hint generation method.}
    \label{tbl:result_llama_nq}
\end{table}
\begin{table}[htb]
	\resizebox{\columnwidth}{!}{%
		\begin{threeparttable}
			[b]
			\begin{tabular}{@{}l|lllllll@{}}
				\toprule Method       & ACC            & EM             & F1            & PR            & RC             & CON            & BERT           \\
				\midrule \multicolumn{8}{c}{\textit{Zero-Shot}} \\
				\midrule BM25                     & 27.51          & 0              & 4.38          & 2.6           & 26.89          & 23.77          & 65.41          \\
				Contriever               & 8.22           & 0              & 2.42          & 1.37          & 21.7           & 14.12          & 62.41          \\
				DPR                      & 26.53          & 0              & 4.8           & 2.79          & 31.7           & 26.57          & 65.63          \\
				MSS                      & 24.9           & 0              & 4.06          & 2.39          & 27.1           & 21.75          & 64.54          \\
				MSS-DPR                  & 30.17          & 0              & 5.08          & 2.98          & 31.42          & 27.36          & 66             \\
				\midrule LLaMA-70b & 45.13          & 0              & 6.16          & 3.65          & \textbf{44.39} & \textbf{47.39} & 67.05          \\
                \midrule HiGen-Va\tnote{1}         & 52.95          & 0              & 5.83          & 3.42          & 38.15          & 40.26          & 67.37          \\
				HiGen-FT\tnote{1}        & \textbf{54.08} & 0              & \textbf{7.01} & \textbf{4.14} & 40.04          & 45.23          & \textbf{68.79} \\
				\midrule \multicolumn{8}{c}{\textit{Few-Shot}}  \\
				\midrule BM25                     & 35.33          & 11.42          & 22.7          & 22.8          & 32.42          & 31.55          & 73.04          \\
				Contriever               & 17.47          & 5.41           & 10.3          & 9.86          & 18.43          & 13.44          & 66.86          \\
				DPR                      & 30.41          & 9.5            & 20.7          & 20.4          & 30.07          & 29.18          & 72.06          \\
				MSS                      & 28.54          & 9.4            & 18.6          & 18.6          & 26.93          & 25.1           & 70.94          \\
				MSS-DPR                  & 33.51          & 10.29          & 22.1          & 22.2          & 31.51          & 32.23          & 72.9           \\
				\midrule LLaMA-70b & 48.03          & 8.46           & 16.5          & 16.8          & 21.79          & 22.49          & 68.5           \\
                \midrule HiGen-Va\tnote{2}         & 55.87          & 17.52          & 32.1          & 32.1          & 44.22          & 44.88          & 76.87          \\
				HiGen-FT\tnote{2}        & \textbf{56.55} & \textbf{20.28} & \textbf{35.4} & \textbf{35.3} & \textbf{47.32} & \textbf{49.9}  & \textbf{78.51}  \\
				\bottomrule
			\end{tabular}
			\begin{tablenotes}
				\item[1] 10 hints, Convergence reranking. \item[2] 7 hints, Convergence
				reranking.
			\end{tablenotes}
		\end{threeparttable}%
	}
    \caption{The results for \textbf{LLaMA-7b} used as the reader on \textbf{WebQ}, utilizing zero-shot and few-shot strategies.The footnotes provide information on the optimal number of hints and the ranking method chosen to achieve the best results for each learning strategy and hint generation method.}
    \label{tbl:result_llama_webq}
\end{table}

\begin{table*}[htb]
	\centering
	\resizebox{\textwidth}{!}{%
	\begin{tabular}{@{}lllllllllll@{}}
		\toprule
		\multicolumn{1}{l|}{Hint Generator} & \# of Params & \# of Hints & Ranking & ACC           & EM            & F1             & PR             & RC             & CON           & BERT                         \\ \midrule
		\multicolumn{11}{c}{\textit{Zero-Shot}}                                                                                                                                                        \\ \midrule
		\multicolumn{1}{l|}{LLaMA-FT~\citep{mozafari2024triviahg}}       & 7b           & 2           & T5      & 79.0          & 0             & 11.39          & 6.75           & 54.58          & 74.0          & 71.87                        \\
		\multicolumn{1}{l|}{LLaMA-Va~\citep{touvron2023llama}}       & 7b           & 2           & Conv    & 68.0          & 0             & 9.37           & 5.57           & 47.48          & 65.0          & 70.52                        \\
		\multicolumn{1}{l|}{LLaMA-Va~\citep{touvron2023llama}}       & 13b          & 2           & T5      & 79.0          & 0             & 9.98           & 5.91           & 48.09          & 73.0          & 71.39                        \\
		\multicolumn{1}{l|}{LLaMA-FT~\citep{mozafari2024triviahg}}       & 13b          & 5           & Conv    & 83.0          & 0             & 10.65          & 6.31           & 48.23          & 76.0          & 71.68                        \\
		\multicolumn{1}{l|}{LLaMA-Va~\citep{touvron2023llama}}       & 70b          & 2           & Def    & 78.0          & 0             & 10.12          & 5.84           & 53.18          & 77.0          & 71.44                        \\
		\multicolumn{1}{l|}{WizardLM~\citep{xu2024wizardlm}}       & 70b          & 5           & T5      & 80.0          & 0             & 9.94           & 5.9            & 47.22          & 75.0          & 71.58                        \\
		\multicolumn{1}{l|}{GPT 3.5~\citep{brown2020language}}        & 175b         & 2           & Conv    & 81.0          & 0             & 10.6           & 6.13           & \textbf{59.37} & 81.0          & 71.45                        \\
		\multicolumn{1}{l|}{Gemini~\citep{team2023gemini}}         & -         & 7           & Def     & 88.0          & 0             & 11.83          & 7.05           & 53             & 88.0          & 72.47                        \\
		\multicolumn{1}{l|}{LLaMA-FT~\citep{mozafari2024triviahg}}       & 70b          & 2           & Conv    & 83.0          & 0             & 11.28          & 6.77           & 50.72          & 81.0          & 72.41                        \\
		\multicolumn{1}{l|}{Copilot}        & -        & 7           & T5      & 92.0          & 0             & \textbf{11.89} & \textbf{7.09}  & 55.32          & \textbf{90.0} & 72.69                        \\
		\multicolumn{1}{l|}{GPT 4~\citep{achiam2023gpt}}          & -        & 5           & Def     & \textbf{96.0} & 0             & 11.5           & 6.8            & 53.97          & 89.0          & \textbf{73.2}                \\ \midrule
		\multicolumn{11}{c}{\textit{Few-Shot}}                                                                                                                                                         \\ \midrule
		\multicolumn{1}{l|}{LLaMA-FT~\citep{mozafari2024triviahg}}       & 7b           & 5           & Conv    & 76.0          & 67.0          & 72.91          & 71.56          & 76.17          & 78.0          & 92.66                        \\
		\multicolumn{1}{l|}{LLaMA-Va~\citep{touvron2023llama}}       & 7b           & 7           & T5      & 76.0          & 57.0          & 67.23          & 65.56          & 71.74          & 72.0          & 90.65                        \\
		\multicolumn{1}{l|}{LLaMA-Va~\citep{touvron2023llama}}       & 13b          & 7           & T5      & 83.0          & 67.0          & 77.04          & 74.87          & 82.17          & 83.0          & 93.33                        \\
		\multicolumn{1}{l|}{LLaMA-FT~\citep{mozafari2024triviahg}}       & 13b          & 10          & Def     & 84.0          & 67.0          & 74.37          & 72.85          & 78.37          & 82.0          & 92.26                        \\
		\multicolumn{1}{l|}{LLaMA-Va~\citep{touvron2023llama}}       & 70b          & 7           & Conv    & 84.0          & 67.0          & 74.29          & 73.18          & 78.87          & 79.0          & 92.09                        \\
		\multicolumn{1}{l|}{WizardLM~\citep{xu2024wizardlm}}       & 70b          & 10          & T5      & 87.0          & 72.0          & 80.04          & 78.29          & 85.17          & 86.0          & 93.67                        \\
		\multicolumn{1}{l|}{GPT 3.5~\citep{brown2020language}}        & 175b         & 7           & Conv    & 88.0          & 72.0          & 79.74          & 78.14          & 83.7           & 84.0          & 93.57                        \\
		\multicolumn{1}{l|}{Gemini~\citep{team2023gemini}}         & -         & 7           & Def     & 90.0          & 73.0          & 81.24          & 79.73          & 85.5           & 89.0          & 94.58                        \\
		\multicolumn{1}{l|}{LLaMA-FT~\citep{mozafari2024triviahg}}       & 70b          & 5           & Def     & 91.0          & 69.0          & 80.06          & 78.11          & 85.87          & 87.0          & 94.02                        \\
		\multicolumn{1}{l|}{Copilot}        & -        & 7           & Conv    & 91.0          & \textbf{77.0} & 86.16          & 84.07          & 92             & \textbf{94.0} & \textbf{95.57}               \\
		\multicolumn{1}{l|}{GPT 4~\citep{achiam2023gpt}}          & -        & 10          & Def     & \textbf{93.0} & 76.0          & \textbf{87.29} & \textbf{85.03} & \textbf{92.17} & 92.0          & 95.46                        \\ \bottomrule
	\end{tabular}%
 }
	\caption{The results of LLaMA-7b as the core of the \method system across different LLMs, generating hints for 100 questions. More detailed results are given in a series of  tables ranging from Table~\ref{tbl:ablation_study_llama_7b_ft} to Table~\ref{tbl:ablation_study_3b} in Appendix~\ref{apx:additional_experiments}.}
	\label{tbl:ablation_study_llama}
\end{table*}

\begin{table}[htb]
	\resizebox{\columnwidth}{!}{%
		\begin{threeparttable}
			[b]
			\begin{tabular}{@{}lllllll@{}}
				\toprule \multicolumn{1}{c|}{\multirow{2}{*}{Method}}         & \multicolumn{3}{c|}{TriviaQA} & \multicolumn{3}{c}{NQ} \\
				\multicolumn{1}{c|}{}                 & EM            & RC             & \multicolumn{1}{l|}{CON}            & EM            & RC          & CON            \\
				\midrule \multicolumn{7}{c}{\textit{Without using rerankers}} \\
				\midrule \multicolumn{1}{l|}{BM25}    & 38.15         & 52.8           & \multicolumn{1}{l|}{51.06}          & 10.33         & 23.28       & 19.14          \\
				\multicolumn{1}{l|}{Contriever}       & 33.54         & 47.31          & \multicolumn{1}{l|}{42.86}          & 6.84          & 16.17       & 11.19          \\
				\multicolumn{1}{l|}{DPR}              & 37.15         & 51.06          & \multicolumn{1}{l|}{49.16}          & 7.15          & 16.92       & 11.63          \\
				\multicolumn{1}{l|}{MSS}              & 33.99         & 47.84          & \multicolumn{1}{l|}{43.41}          & 7.15          & 17.2        & 12.05          \\
				\midrule \multicolumn{7}{c}{\textit{With using rerankers}}    \\
				\midrule \multicolumn{1}{l|}{MSS+UPR} & 53.1          & 67.3           & \multicolumn{1}{l|}{60.6}           & 25.4          & 40.7        & 31             \\
				\multicolumn{1}{l|}{DPR+UPR}          & 53.9          & 68.7           & \multicolumn{1}{l|}{62}             & \textbf{25.6} & \textbf{42} & 33.1           \\
				\midrule \multicolumn{7}{c}{\textit{Our method}}              \\
				\midrule \multicolumn{1}{l|}{HiGen-Va} & 50.76         & 65.12          & \multicolumn{1}{l|}{65.92}          & 18.48         & 34.58       & 33.24          \\
				\multicolumn{1}{l|}{HiGen-FT}         & \textbf{54.62} & \textbf{69.53} & \multicolumn{1}{l|}{\textbf{70.15}} & 20.72         & 37.19       & \textbf{36.81} \\
				\bottomrule
			\end{tabular}
		\end{threeparttable}%
	}
    \caption{Comparison of reults between baselines without rerankers, baselines with rerankers, and \method.}
    \label{tbl:ablation_study_reranker}
\end{table}

\begin{figure*}[tb]
	\includegraphics[width=\textwidth]{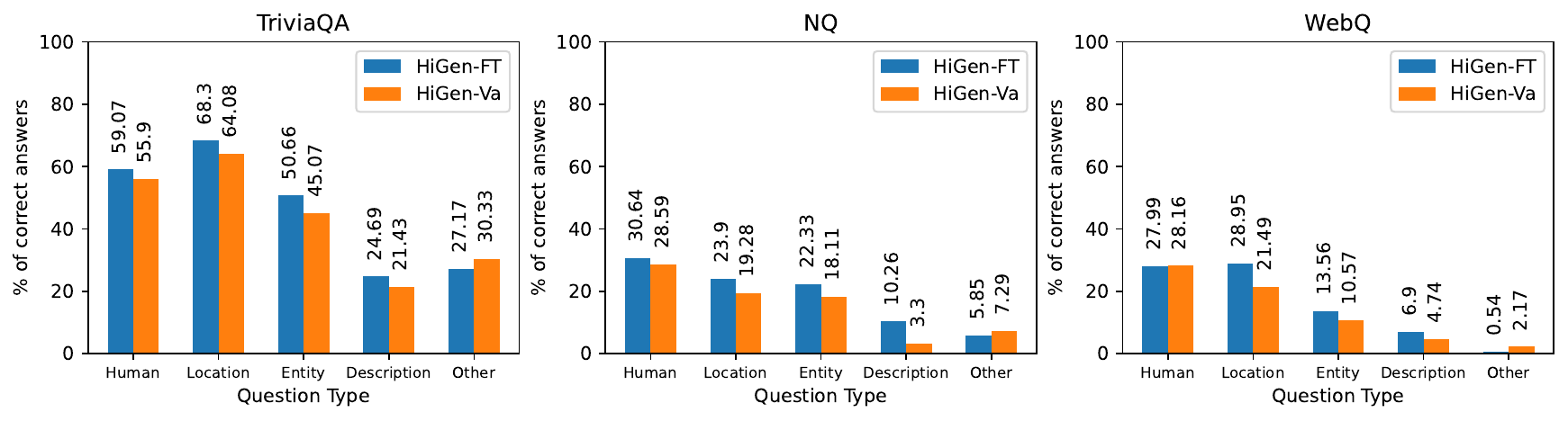}
	\caption{Exact Match values for TriviaQA, NQ, and WebQ datasets categorized by
		question type, based on the optimal settings for both HiGen-Va and HiGen-FT using few-shot learning on LLaMA-7b.}
	\label{fig:experiments}
\end{figure*}

\section{Results}\label{s:results}

\subsection{Context Length}\label{ss:context_length}
We first discuss the average lengths of contexts retrieved or generated, based on the number of words, by different models within the Context-Preparator component. As noted in Section~\ref{s:introduction}, our approach yields contexts that are longer than those produced by generation-based methods but shorter than those from retrieval-based methods. Table~\ref{tbl:avg_length_hint_gen_ret} provides details on the average lengths of hints, generated contexts, and retrieved passages across the TriviaQA, NQ, and WebQ datasets. The data indicates that the length of hints produced by both \textbf{HiGen-FT} and \textbf{HiGen-Va} methods are shorter than those from all retrieval-based methods. However, when compared with \textbf{LLaMA-70b} used as a generative approach, the hints are longer.

\subsection{Results of \method}\label{ss:results_on_qa}

In this section, we present and analyze the performance and results of the \method approach, comparing it against various baselines. As previously mentioned, our experimental framework encompasses a range of setups, including different datasets (Section~\ref{ss:dataset}), baseline models (Section~\ref{ss:baselines}), HG systems, orders of hints, numbers of hints (Section~\ref{ss:hint_generation_methods}), and readers (Section~\ref{ss:readers}). This comprehensive evaluation helps in assessing the robustness and effectiveness of the \method approach across multiple dimensions.

Table~\ref{tbl:result_t5} presents the performance of the T5-3b model as the Reader component, utilizing zero-shot and few-shot learning strategies across the specified datasets, measured by Exact Match and F1 scores. The results indicate that \textbf{HiGen-FT} achieves the best performance in both learning strategies. Additionally, the outcomes from the few-shot learning strategy surpass those of the zero-shot learning strategy. For a more detailed analysis of T5-3b's performance using HiGen-Va on TriviaQA, NQ, and WebQ datasets, readers can refer to Table~\ref{tbl:3b-va-trivia} to Table~\ref{tbl:3b-va-webq} in Appendix~\ref{apx:additional_experiments}. Tables~\ref{tbl:3b-ft-trivia} to Table~\ref{tbl:3b-ft-webq} provide information on T5-3b's performance using HiGen-FT for these datasets.

Table~\ref{tbl:result_llama_trivia}, Table~\ref{tbl:result_llama_nq}, and Table~\ref{tbl:result_llama_webq} show the performance of the LLaMA-7b model as the Reader component across different experimental setups. The results show that in the few-shot strategy, \textbf{HiGen-FT} consistently delivers the best outcomes across all three datasets. However, the performance under the zero-shot learning strategy varies. The EM metric is zero in the zero-shot learning strategy because the final generated answers are often text passages rather than short answers.

For the TriviaQA dataset, \textbf{LLaMA-70b} leads in F1, Precision, and Recall metrics. In the case of the NQ dataset, \textbf{LLaMA-70b} performs best regarding Recall, while for the WebQ dataset, \textbf{LLaMA-70b} excels in both Recall and Contains metrics. For other metrics across these datasets, the \textbf{\method} approach outperforms the rest. 

Figure~\ref{fig:experiments} displays Exact Match scores for the TriviaQA, NQ, and WebQ datasets, broken down by the question type, under the optimal settings for both HiGen-Va and HiGen-FT using few-shot strategy on LLaMA-7b. The figure illustrates that \textbf{HiGen-FT} outperforms HiGen-Va across various question types. For more detailed analysis of LLaMA-7b's performance using HiGen-Va and HiGen-FT on TriviaQA, NQ, and WebQ datasets, readers can refer to tables from Table~\ref{tbl:llama-va-trivia} to Table~\ref{tbl:llama-ft-webq} in Appendix~\ref{apx:additional_experiments}.

In Appendix~\ref{apx:case_study}, Table~\ref{tbl:case_study_trivia_nq_webq} presents a comparison of answers for a random selection of questions from the TriviaQA, NQ, and WebQ datasets, using DPR, LLaMA-70b, and \method. Table~\ref{tbl:case_study_retriever_genrative_higenqa_zero} shows the answers generated from contexts retrieved by MSS-DPR, contexts generated by LLaMA-70b, and hints generated by \method using the LLaMA-7b model in a zero-shot learning strategy. Table~\ref{tbl:case_study_retriever}, Table~\ref{tbl:case_study_generative}, and Table~\ref{tbl:case_study_higenqa} illustrate the answers generated under a few-shot learning strategy by MSS-DPR, LLaMA-70b, and \method, respectively, using the LLaMA-7b model.

\subsection{Ablation Study} \label{ss:ablation_study}

\paragraph{Impact of various LLMs} We investigate the impact of LLMs used as the core in the HG method, producing various hints for some random questions from the TriviaQA dataset. Utilizing various LLMs, we generate hints per each question. Table~\ref{tbl:ablation_study_llama} presents the top-performing results for these LLMs as the core of the HG method across different numbers of hints and reranking methods, with LLaMA-7b serving as the Reader. The findings reveal that \textbf{Copilot} and \textbf{GPT-4}~\citep{achiam2023gpt} deliver the best performance for zero-shot and few-shot learning strategies, respectively, highlighting that a more knowledgeable core can produce higher-quality hints. 
The results for T5-3b as the Reader component are given 
in Table~\ref{tbl:ablation_study_3b} in Appendix~\ref{apx:additional_experiments}.

\paragraph{Impact of Rerankers} Finally, we evaluate the impact of rerankers on retrieval-based methods and the \method approach to determine how \method performs relative to other methods when rerankers are used. Table~\ref{tbl:ablation_study_reranker} displays the results for retrievers without rerankers, with the UPR-reranker~\citep{sachan-etal-2022-improving}, and \method for both the TriviaQA and NQ datasets. The results show that \method surpasses others on TriviaQA dataset. Yet, while \method achieves the best results with the Contains metric for the NQ dataset, UPR-reranker performs better in other metrics.
\section{Conclusion} \label{s:conclusion}
In this paper, we introduced a novel approach to the Context-Preparator in QA systems called \method that generates hints instead of relying on retrieved passages or generated contexts. To thoroughly test its effectiveness, we designed a variety of experimental setups, aiming to cover a broad spectrum of possible scenarios. Our findings reveal that this new approach consistently surpasses traditional baseline methods, including both retrieval-based and generation-based approaches, on the TriviaQA, NQ, and WebQ datasets across multiple evaluation metrics. Moreover, we demonstrated that different configurations, such as employing various LLMs as the core of the HG method and adjusting ranking methods and the number of hints, significantly boost the performance of our approach. Our future work will focus on more complex questions such as multi-hop questions \cite{DBLP:journals/ftir/MaviJJ24} requiring comprehensive reasoning.
\section*{Limitations}
Our study has the following limitations:
\begin{itemize}
    \item The \method approach leverages the capabilities of LLMs to generate high-quality hints by utilizing the extensive knowledge stored within these models. However, a key limitation is that the hints are based on outdated data, as they reflect only the information available up to the LLMs' last training period. This could lead to less relevant and accurate hints, especially in rapidly evolving fields like technology, medicine, and science.
    \item The computation of HICOS scores using LLMs is both time-consuming and resource-intensive, presenting significant challenges. Although sorting hints by descending HICOS scores delivers optimal performance, the process is complex and demands substantial computational resources. This could limit the use of such systems in scenarios that require fast or cost-effective solutions, particularly in environments with limited hardware capabilities or where minimizing operational costs is essential.
    \item The LLMs in the reader component were intentionally not fine-tuned on the TriviaQA, NQ, and WebQ datasets to assess the Hint Generation (HG) method's effectiveness as a Context-Preparator without introducing bias. This approach ensures that the evaluation focuses solely on how well the HG method enhances the reader's performance by preparing context, rather than relying on any pre-existing familiarity with the dataset content.
    \item We have focused in this work on factoid questions since the current hint generation systems, the same as many QA systems, are typically designed to work only with factoid questions, and we rely on these systems to generate hints in our method.
\end{itemize}
\section*{Ethical Considerations} \label{s:ethics}

Our study employs the GPT models, governed by the OpenAI License and Apache-2.0 license, and the LLaMA model, distributed under Meta's LLaMA 2 Community License Agreement. We adhere to these licenses for all applications. Moreover, the datasets we use are sourced from repositories authorized for academic purposes. The artifacts developed during our research are released under the MIT license to promote easy modification and use by the research community. We have ensured that our data handling, model training, and dissemination of results comply with ethical standards and legal requirements related to each utilized artifact.
\section*{Acknowledgments}
The computational results presented here have been achieved (in part) using the LEO HPC infrastructure of the University of Innsbruck.

% Bibliography entries for the entire Anthology, followed by custom entries
%\bibliography{anthology,custom}
% Custom bibliography entries only

\bibliography{custom}

\begin{thebibliography}{49}
\providecommand{\natexlab}[1]{#1}

\bibitem[{Abdel-Nabi et~al.(2023)Abdel-Nabi, Awajan, and Ali}]{Abdel-Nabi2023}
Heba Abdel-Nabi, Arafat Awajan, and Mostafa~Z. Ali. 2023.
\newblock \href {https://doi.org/10.1007/s10115-022-01783-5} {Deep learning-based question answering: a survey}.
\newblock \emph{Knowledge and Information Systems}, 65(4):1399--1485.

\bibitem[{Achiam et~al.(2023)Achiam, Adler, Agarwal, Ahmad, Akkaya, Aleman, Almeida, Altenschmidt, Altman, Anadkat et~al.}]{achiam2023gpt}
Josh Achiam, Steven Adler, Sandhini Agarwal, Lama Ahmad, Ilge Akkaya, Florencia~Leoni Aleman, Diogo Almeida, Janko Altenschmidt, Sam Altman, Shyamal Anadkat, et~al. 2023.
\newblock \href {https://doi.org//10.48550/arXiv.2303.08774} {Gpt-4 technical report}.
\newblock \emph{arXiv preprint arXiv:2303.08774}.

\bibitem[{Berant et~al.(2013)Berant, Chou, Frostig, and Liang}]{berant-etal-2013-semantic}
Jonathan Berant, Andrew Chou, Roy Frostig, and Percy Liang. 2013.
\newblock \href {https://aclanthology.org/D13-1160} {Semantic parsing on {F}reebase from question-answer pairs}.
\newblock In \emph{Proceedings of the 2013 Conference on Empirical Methods in Natural Language Processing}, pages 1533--1544, Seattle, Washington, USA. Association for Computational Linguistics.

\bibitem[{Bevilacqua et~al.(2022)Bevilacqua, Ottaviano, Lewis, Yih, Riedel, and Petroni}]{bevilacqua2022autoregressive}
Michele Bevilacqua, Giuseppe Ottaviano, Patrick Lewis, Scott Yih, Sebastian Riedel, and Fabio Petroni. 2022.
\newblock \href {https://proceedings.neurips.cc/paper_files/paper/2022/file/cd88d62a2063fdaf7ce6f9068fb15dcd-Paper-Conference.pdf} {Autoregressive search engines: Generating substrings as document identifiers}.
\newblock In \emph{Advances in Neural Information Processing Systems}, volume~35, pages 31668--31683. Curran Associates, Inc.

\bibitem[{Bollacker et~al.(2008)Bollacker, Evans, Paritosh, Sturge, and Taylor}]{10.1145/1376616.1376746}
Kurt Bollacker, Colin Evans, Praveen Paritosh, Tim Sturge, and Jamie Taylor. 2008.
\newblock \href {https://doi.org/10.1145/1376616.1376746} {Freebase: a collaboratively created graph database for structuring human knowledge}.
\newblock In \emph{Proceedings of the 2008 ACM SIGMOD International Conference on Management of Data}, SIGMOD '08, page 1247–1250, New York, NY, USA. Association for Computing Machinery.

\bibitem[{Brown et~al.(2020)Brown, Mann, Ryder, Subbiah, Kaplan, Dhariwal, Neelakantan, Shyam, and Others}]{brown2020language}
Tom~B. Brown, Benjamin Mann, Nick Ryder, Melanie Subbiah, Jared Kaplan, Prafulla Dhariwal, Arvind Neelakantan, Pranav Shyam, and Others. 2020.
\newblock \href {https://dl.acm.org/doi/abs/10.5555/3495724.3495883} {Language models are few-shot learners}.
\newblock In \emph{Proceedings of the 34th International Conference on Neural Information Processing Systems}, NIPS '20, Red Hook, NY, USA. Curran Associates Inc.

\bibitem[{Chowdhery et~al.(2024)Chowdhery, Narang, Devlin, Bosma, Mishra, Roberts, Barham, Chung, Sutton, Gehrmann, Schuh, and Others}]{chowdhery2023palm}
Aakanksha Chowdhery, Sharan Narang, Jacob Devlin, Maarten Bosma, Gaurav Mishra, Adam Roberts, Paul Barham, Hyung~Won Chung, Charles Sutton, Sebastian Gehrmann, Parker Schuh, and Others. 2024.
\newblock \href {https://dl.acm.org/doi/10.5555/3648699.3648939} {Palm: scaling language modeling with pathways}.
\newblock \emph{J. Mach. Learn. Res.}, 24(1).

\bibitem[{Devlin et~al.(2019)Devlin, Chang, Lee, and Toutanova}]{devlin-etal-2019-bert}
Jacob Devlin, Ming-Wei Chang, Kenton Lee, and Kristina Toutanova. 2019.
\newblock \href {https://doi.org/10.18653/v1/N19-1423} {{BERT}: Pre-training of deep bidirectional transformers for language understanding}.
\newblock In \emph{Proceedings of the 2019 Conference of the North {A}merican Chapter of the Association for Computational Linguistics: Human Language Technologies, Volume 1 (Long and Short Papers)}, pages 4171--4186, Minneapolis, Minnesota. Association for Computational Linguistics.

\bibitem[{Formal et~al.(2021)Formal, Piwowarski, and Clinchant}]{formal2021splade}
Thibault Formal, Benjamin Piwowarski, and St\'{e}phane Clinchant. 2021.
\newblock \href {https://doi.org/10.1145/3404835.3463098} {Splade: Sparse lexical and expansion model for first stage ranking}.
\newblock In \emph{Proceedings of the 44th International ACM SIGIR Conference on Research and Development in Information Retrieval}, SIGIR '21, page 2288–2292, New York, NY, USA. Association for Computing Machinery.

\bibitem[{Gysel et~al.(2018)Gysel, de~Rijke, and Kanoulas}]{10.1145/3196826}
Christophe~Van Gysel, Maarten de~Rijke, and Evangelos Kanoulas. 2018.
\newblock \href {https://doi.org/10.1145/3196826} {Neural vector spaces for unsupervised information retrieval}.
\newblock \emph{ACM Trans. Inf. Syst.}, 36(4).

\bibitem[{Izacard et~al.(2022)Izacard, Caron, Hosseini, Riedel, Bojanowski, Joulin, and Grave}]{izacard2022unsupervised}
Gautier Izacard, Mathilde Caron, Lucas Hosseini, Sebastian Riedel, Piotr Bojanowski, Armand Joulin, and Edouard Grave. 2022.
\newblock \href {https://openreview.net/forum?id=jKN1pXi7b0} {Unsupervised dense information retrieval with contrastive learning}.
\newblock \emph{Transactions on Machine Learning Research}.

\bibitem[{{Jangra} et~al.(2024){Jangra}, {Mozafari}, {Jatowt}, and {Muresan}}]{jangra2024navigating}
Anubhav {Jangra}, Jamshid {Mozafari}, Adam {Jatowt}, and Smaranda {Muresan}. 2024.
\newblock \href {https://doi.org/10.48550/arXiv.2404.04728} {{Navigating the Landscape of Hint Generation Research: From the Past to the Future}}.
\newblock \emph{arXiv e-prints}, arXiv:2404.04728.

\bibitem[{Jatowt et~al.(2023)Jatowt, Gehrer, and F\"{a}rber}]{10.1145/3578337.3605119}
Adam Jatowt, Calvin Gehrer, and Michael F\"{a}rber. 2023.
\newblock \href {https://doi.org/10.1145/3578337.3605119} {Automatic hint generation}.
\newblock In \emph{Proceedings of the 2023 ACM SIGIR International Conference on Theory of Information Retrieval}, ICTIR '23, page 117–123, New York, NY, USA. Association for Computing Machinery.

\bibitem[{Jin et~al.(2012)Jin, Barnes, Stamper, Eagle, Johnson, and Lehmann}]{10.1007/978-3-642-30950-2_40}
Wei Jin, Tiffany Barnes, John Stamper, Michael~John Eagle, Matthew~W. Johnson, and Lorrie Lehmann. 2012.
\newblock \href {https://doi.org/10.1007/978-3-642-30950-2_40} {Program representation for automatic hint generation for a data-driven novice programming tutor}.
\newblock In \emph{Proceedings of the 11th International Conference on Intelligent Tutoring Systems}, ITS'12, page 304–309, Berlin, Heidelberg. Springer-Verlag.

\bibitem[{Joshi et~al.(2017)Joshi, Choi, Weld, and Zettlemoyer}]{joshi-etal-2017-triviaqa}
Mandar Joshi, Eunsol Choi, Daniel Weld, and Luke Zettlemoyer. 2017.
\newblock \href {https://doi.org/10.18653/v1/P17-1147} {{T}rivia{QA}: A large scale distantly supervised challenge dataset for reading comprehension}.
\newblock In \emph{Proceedings of the 55th Annual Meeting of the Association for Computational Linguistics (Volume 1: Long Papers)}, pages 1601--1611, Vancouver, Canada. Association for Computational Linguistics.

\bibitem[{Kamalloo et~al.(2023)Kamalloo, Dziri, Clarke, and Rafiei}]{kamalloo-etal-2023-evaluating}
Ehsan Kamalloo, Nouha Dziri, Charles Clarke, and Davood Rafiei. 2023.
\newblock \href {https://doi.org/10.18653/v1/2023.acl-long.307} {Evaluating open-domain question answering in the era of large language models}.
\newblock In \emph{Proceedings of the 61st Annual Meeting of the Association for Computational Linguistics (Volume 1: Long Papers)}, pages 5591--5606, Toronto, Canada. Association for Computational Linguistics.

\bibitem[{Karpukhin et~al.(2020)Karpukhin, Oguz, Min, Lewis, Wu, Edunov, Chen, and Yih}]{karpukhin-etal-2020-dense}
Vladimir Karpukhin, Barlas Oguz, Sewon Min, Patrick Lewis, Ledell Wu, Sergey Edunov, Danqi Chen, and Wen-tau Yih. 2020.
\newblock \href {https://doi.org/10.18653/v1/2020.emnlp-main.550} {Dense passage retrieval for open-domain question answering}.
\newblock In \emph{Proceedings of the 2020 Conference on Empirical Methods in Natural Language Processing (EMNLP)}, pages 6769--6781, Online. Association for Computational Linguistics.

\bibitem[{Kwiatkowski et~al.(2019)Kwiatkowski, Palomaki, Redfield, Collins, Parikh, Alberti, Epstein, Polosukhin, Devlin, Lee, Toutanova, Jones, Kelcey, Chang, Dai, Uszkoreit, Le, and Petrov}]{kwiatkowski-etal-2019-natural}
Tom Kwiatkowski, Jennimaria Palomaki, Olivia Redfield, Michael Collins, Ankur Parikh, Chris Alberti, Danielle Epstein, Illia Polosukhin, Jacob Devlin, Kenton Lee, Kristina Toutanova, Llion Jones, Matthew Kelcey, Ming-Wei Chang, Andrew~M. Dai, Jakob Uszkoreit, Quoc Le, and Slav Petrov. 2019.
\newblock \href {https://doi.org/10.1162/tacl_a_00276} {Natural questions: A benchmark for question answering research}.
\newblock \emph{Transactions of the Association for Computational Linguistics}, 7:452--466.

\bibitem[{Lewis et~al.(2020)Lewis, Liu, Goyal, Ghazvininejad, Mohamed, Levy, Stoyanov, and Zettlemoyer}]{lewis-etal-2020-bart}
Mike Lewis, Yinhan Liu, Naman Goyal, Marjan Ghazvininejad, Abdelrahman Mohamed, Omer Levy, Veselin Stoyanov, and Luke Zettlemoyer. 2020.
\newblock \href {https://doi.org/10.18653/v1/2020.acl-main.703} {{BART}: Denoising sequence-to-sequence pre-training for natural language generation, translation, and comprehension}.
\newblock In \emph{Proceedings of the 58th Annual Meeting of the Association for Computational Linguistics}, pages 7871--7880, Online. Association for Computational Linguistics.

\bibitem[{{Li} et~al.(2024){Li}, {Jin}, {Zhou}, {Zhang}, {Zhang}, {Zhu}, and {Dou}}]{li2024matching}
Xiaoxi {Li}, Jiajie {Jin}, Yujia {Zhou}, Yuyao {Zhang}, Peitian {Zhang}, Yutao {Zhu}, and Zhicheng {Dou}. 2024.
\newblock \href {https://doi.org/10.48550/arXiv.2404.14851} {{From Matching to Generation: A Survey on Generative Information Retrieval}}.
\newblock \emph{arXiv e-prints}, arXiv:2404.14851.

\bibitem[{{Lin} and {Ma}(2021)}]{lin2021few}
Jimmy {Lin} and Xueguang {Ma}. 2021.
\newblock \href {https://doi.org/10.48550/arXiv.2106.14807} {{A Few Brief Notes on DeepImpact, COIL, and a Conceptual Framework for Information Retrieval Techniques}}.
\newblock \emph{arXiv e-prints}, arXiv:2106.14807.

\bibitem[{Mao et~al.(2021)Mao, He, Liu, Shen, Gao, Han, and Chen}]{mao-etal-2021-reader}
Yuning Mao, Pengcheng He, Xiaodong Liu, Yelong Shen, Jianfeng Gao, Jiawei Han, and Weizhu Chen. 2021.
\newblock \href {https://doi.org/10.18653/v1/2021.findings-acl.29} {Reader-guided passage reranking for open-domain question answering}.
\newblock In \emph{Findings of the Association for Computational Linguistics: ACL-IJCNLP 2021}, pages 344--350, Online. Association for Computational Linguistics.

\bibitem[{Mavi et~al.(2024)Mavi, Jangra, and Jatowt}]{DBLP:journals/ftir/MaviJJ24}
Vaibhav Mavi, Anubhav Jangra, and Adam Jatowt. 2024.
\newblock \href {https://doi.org/10.1561/1500000102} {Multi-hop question answering}.
\newblock \emph{Found. Trends Inf. Retr.}, 17(5):457--586.

\bibitem[{{Mitra} and {Craswell}(2017)}]{mitra2017neural}
Bhaskar {Mitra} and Nick {Craswell}. 2017.
\newblock \href {https://doi.org/10.48550/arXiv.1705.01509} {{Neural Models for Information Retrieval}}.
\newblock \emph{arXiv e-prints}, arXiv:1705.01509.

\bibitem[{Mozafari et~al.(2024)Mozafari, Jangra, and Jatowt}]{mozafari2024triviahg}
Jamshid Mozafari, Anubhav Jangra, and Adam Jatowt. 2024.
\newblock \href {https://doi.org/10.1145/3626772.3657855} {Triviahg: A dataset for automatic hint generation from factoid questions}.
\newblock In \emph{Proceedings of the 47th International ACM SIGIR Conference on Research and Development in Information Retrieval}, SIGIR '24, page 2060–2070, New York, NY, USA. Association for Computing Machinery.

\bibitem[{Oberm\"{u}ller et~al.(2021)Oberm\"{u}ller, Heuer, and Fraser}]{10.1145/3430665.3456344}
Florian Oberm\"{u}ller, Ute Heuer, and Gordon Fraser. 2021.
\newblock \href {https://doi.org/10.1145/3430665.3456344} {Guiding next-step hint generation using automated tests}.
\newblock In \emph{Proceedings of the 26th ACM Conference on Innovation and Technology in Computer Science Education V. 1}, ITiCSE '21, page 220–226, New York, NY, USA. Association for Computing Machinery.

\bibitem[{Ouyang et~al.(2024)Ouyang, Wu, Jiang, Almeida, Wainwright, Mishkin, Zhang, Agarwal, Slama, Ray, Schulman, Hilton, Kelton, Miller, Simens, Askell, Welinder, Christiano, Leike, and Lowe}]{ouyang2022training}
Long Ouyang, Jeff Wu, Xu~Jiang, Diogo Almeida, Carroll~L. Wainwright, Pamela Mishkin, Chong Zhang, Sandhini Agarwal, Katarina Slama, Alex Ray, John Schulman, Jacob Hilton, Fraser Kelton, Luke Miller, Maddie Simens, Amanda Askell, Peter Welinder, Paul Christiano, Jan Leike, and Ryan Lowe. 2024.
\newblock \href {https://dl.acm.org/doi/10.5555/3600270.3602281} {Training language models to follow instructions with human feedback}.
\newblock In \emph{Proceedings of the 36th International Conference on Neural Information Processing Systems}, NIPS '22, Red Hook, NY, USA. Curran Associates Inc.

\bibitem[{Pedregosa et~al.(2011)Pedregosa, Varoquaux, Gramfort, Michel, Thirion, Grisel, Blondel, Prettenhofer, Weiss, Dubourg, Vanderplas, Passos, Cournapeau, Brucher, Perrot, and Duchesnay}]{scikit-learn}
F.~Pedregosa, G.~Varoquaux, A.~Gramfort, V.~Michel, B.~Thirion, O.~Grisel, M.~Blondel, P.~Prettenhofer, R.~Weiss, V.~Dubourg, J.~Vanderplas, A.~Passos, D.~Cournapeau, M.~Brucher, M.~Perrot, and E.~Duchesnay. 2011.
\newblock \href {http://jmlr.org/papers/v12/pedregosa11a.html} {Scikit-learn: Machine learning in {P}ython}.
\newblock \emph{Journal of Machine Learning Research}, 12:2825--2830.

\bibitem[{Raffel et~al.(2020)Raffel, Shazeer, Roberts, Lee, Narang, Matena, Zhou, Li, and Liu}]{10.5555/3455716.3455856}
Colin Raffel, Noam Shazeer, Adam Roberts, Katherine Lee, Sharan Narang, Michael Matena, Yanqi Zhou, Wei Li, and Peter~J. Liu. 2020.
\newblock \href {https://dl.acm.org/doi/10.5555/3455716.3455856} {Exploring the limits of transfer learning with a unified text-to-text transformer}.
\newblock \emph{J. Mach. Learn. Res.}, 21(1).

\bibitem[{Rivers et~al.(2016)Rivers, Harpstead, and Koedinger}]{10.1145/2960310.2960333}
Kelly Rivers, Erik Harpstead, and Ken Koedinger. 2016.
\newblock \href {https://doi.org/10.1145/2960310.2960333} {Learning curve analysis for programming: Which concepts do students struggle with?}
\newblock In \emph{Proceedings of the 2016 ACM Conference on International Computing Education Research}, ICER '16, page 143–151, New York, NY, USA. Association for Computing Machinery.

\bibitem[{Robertson and Zaragoza(2009)}]{robertson2009probabilistic}
Stephen Robertson and Hugo Zaragoza. 2009.
\newblock \href {https://doi.org/10.1561/1500000019} {The probabilistic relevance framework: Bm25 and beyond}.
\newblock \emph{Found. Trends Inf. Retr.}, 3(4):333–389.

\bibitem[{Rogers et~al.(2023)Rogers, Gardner, and Augenstein}]{10.1145/3560260}
Anna Rogers, Matt Gardner, and Isabelle Augenstein. 2023.
\newblock \href {https://doi.org/10.1145/3560260} {Qa dataset explosion: A taxonomy of nlp resources for question answering and reading comprehension}.
\newblock \emph{ACM Comput. Surv.}, 55(10).

\bibitem[{Sachan et~al.(2022)Sachan, Lewis, Joshi, Aghajanyan, Yih, Pineau, and Zettlemoyer}]{sachan-etal-2022-improving}
Devendra Sachan, Mike Lewis, Mandar Joshi, Armen Aghajanyan, Wen-tau Yih, Joelle Pineau, and Luke Zettlemoyer. 2022.
\newblock \href {https://doi.org/10.18653/v1/2022.emnlp-main.249} {Improving passage retrieval with zero-shot question generation}.
\newblock In \emph{Proceedings of the 2022 Conference on Empirical Methods in Natural Language Processing}, pages 3781--3797, Abu Dhabi, United Arab Emirates. Association for Computational Linguistics.

\bibitem[{Sachan et~al.(2021)Sachan, Patwary, Shoeybi, Kant, Ping, Hamilton, and Catanzaro}]{sachan-etal-2021-end}
Devendra Sachan, Mostofa Patwary, Mohammad Shoeybi, Neel Kant, Wei Ping, William~L. Hamilton, and Bryan Catanzaro. 2021.
\newblock \href {https://doi.org/10.18653/v1/2021.acl-long.519} {End-to-end training of neural retrievers for open-domain question answering}.
\newblock In \emph{Proceedings of the 59th Annual Meeting of the Association for Computational Linguistics and the 11th International Joint Conference on Natural Language Processing (Volume 1: Long Papers)}, pages 6648--6662, Online. Association for Computational Linguistics.

\bibitem[{Salton et~al.(1983)Salton, Fox, and Wu}]{salton1983extended}
Gerard Salton, Edward~A. Fox, and Harry Wu. 1983.
\newblock \href {https://doi.org/10.1145/182.358466} {Extended boolean information retrieval}.
\newblock \emph{Commun. ACM}, 26(11):1022–1036.

\bibitem[{Scao et~al.(2022)Scao, Fan, Akiki, Pavlick, Ili{\'c}, Hesslow, Castagn{\'e}, Luccioni, Yvon et~al.}]{workshop2022bloom}
Teven~Le Scao, Angela Fan, Christopher Akiki, Ellie Pavlick, Suzana Ili{\'c}, Daniel Hesslow, Roman Castagn{\'e}, Alexandra~Sasha Luccioni, Fran{\c{c}}ois Yvon, et~al. 2022.
\newblock \href {https://doi.org/10.48550/arXiv.2211.05100} {{BLOOM: A 176B-Parameter Open-Access Multilingual Language Model}}.
\newblock \emph{arXiv e-prints}, arXiv:2211.05100.

\bibitem[{Siddiqui and Tiwary(2005)}]{10.1007/11554028_10}
Tanveer~J. Siddiqui and Uma~Shanker Tiwary. 2005.
\newblock \href {https://doi.org/10.1007/11554028_10} {Integrating relation and keyword matching in information retrieval}.
\newblock In \emph{Proceedings of the 9th International Conference on Knowledge-Based Intelligent Information and Engineering Systems - Volume Part IV}, KES'05, page 64–73, Berlin, Heidelberg. Springer-Verlag.

\bibitem[{Sutskever et~al.(2014)Sutskever, Vinyals, and Le}]{NIPS2014_a14ac55a}
Ilya Sutskever, Oriol Vinyals, and Quoc~V. Le. 2014.
\newblock \href {https://dl.acm.org/doi/10.5555/2969033.2969173} {Sequence to sequence learning with neural networks}.
\newblock In \emph{Proceedings of the 27th International Conference on Neural Information Processing Systems - Volume 2}, NIPS'14, page 3104–3112, Cambridge, MA, USA. MIT Press.

\bibitem[{Tay et~al.(2024)Tay, Tran, Dehghani, Ni, Bahri, Mehta, Qin, Hui, Zhao, Gupta, Schuster, Cohen, and Metzler}]{tay2022transformer}
Yi~Tay, Vinh~Q. Tran, Mostafa Dehghani, Jianmo Ni, Dara Bahri, Harsh Mehta, Zhen Qin, Kai Hui, Zhe Zhao, Jai Gupta, Tal Schuster, William~W. Cohen, and Donald Metzler. 2024.
\newblock \href {https://dl.acm.org/doi/10.5555/3600270.3601857} {Transformer memory as a differentiable search index}.
\newblock In \emph{Proceedings of the 36th International Conference on Neural Information Processing Systems}, NIPS '22, Red Hook, NY, USA. Curran Associates Inc.

\bibitem[{Team et~al.(2023)Team, Anil, Borgeaud, Wu, Alayrac, Yu, Soricut, Schalkwyk, Dai, Hauth et~al.}]{team2023gemini}
Gemini Team, Rohan Anil, Sebastian Borgeaud, Yonghui Wu, Jean-Baptiste Alayrac, Jiahui Yu, Radu Soricut, Johan Schalkwyk, Andrew~M Dai, Anja Hauth, et~al. 2023.
\newblock \href {https://doi.org/10.48550/arXiv.2312.11805} {{Gemini: A Family of Highly Capable Multimodal Models}}.
\newblock \emph{arXiv e-prints}, arXiv:2312.11805.

\bibitem[{{Touvron} et~al.(2023){Touvron}, {Lavril}, {Izacard}, {Martinet}, {Lachaux}, {Lacroix}, {Rozi{\`e}re}, {Goyal}, {Hambro}, {Azhar}, {Rodriguez}, {Joulin}, {Grave}, and {Lample}}]{touvron2023llama}
Hugo {Touvron}, Thibaut {Lavril}, Gautier {Izacard}, Xavier {Martinet}, Marie-Anne {Lachaux}, Timoth{\'e}e {Lacroix}, Baptiste {Rozi{\`e}re}, Naman {Goyal}, Eric {Hambro}, Faisal {Azhar}, Aurelien {Rodriguez}, Armand {Joulin}, Edouard {Grave}, and Guillaume {Lample}. 2023.
\newblock \href {https://doi.org/10.48550/arXiv.2302.13971} {{LLaMA: Open and Efficient Foundation Language Models}}.
\newblock \emph{arXiv e-prints}, arXiv:2302.13971.

\bibitem[{{Wang} et~al.(2022){Wang}, {Yang}, {Huang}, {Jiao}, {Yang}, {Jiang}, {Majumder}, and {Wei}}]{wang2022text}
Liang {Wang}, Nan {Yang}, Xiaolong {Huang}, Binxing {Jiao}, Linjun {Yang}, Daxin {Jiang}, Rangan {Majumder}, and Furu {Wei}. 2022.
\newblock \href {https://doi.org/10.48550/arXiv.2212.03533} {{Text Embeddings by Weakly-Supervised Contrastive Pre-training}}.
\newblock \emph{arXiv e-prints}, arXiv:2212.03533.

\bibitem[{Wang et~al.(2023)Wang, Yang, Huang, Jiao, Yang, Jiang, Majumder, and Wei}]{wang-etal-2023-simlm}
Liang Wang, Nan Yang, Xiaolong Huang, Binxing Jiao, Linjun Yang, Daxin Jiang, Rangan Majumder, and Furu Wei. 2023.
\newblock \href {https://doi.org/10.18653/v1/2023.acl-long.125} {{S}im{LM}: Pre-training with representation bottleneck for dense passage retrieval}.
\newblock In \emph{Proceedings of the 61st Annual Meeting of the Association for Computational Linguistics (Volume 1: Long Papers)}, pages 2244--2258, Toronto, Canada. Association for Computational Linguistics.

\bibitem[{Wang et~al.(2024)Wang, Hou, Wang, Miao, Wu, Sun, Chen, Xia, Chi, Zhao, Liu, Xie, Sun, Deng, Zhang, and Yang}]{wang2022neural}
Yujing Wang, Yingyan Hou, Haonan Wang, Ziming Miao, Shibin Wu, Hao Sun, Qi~Chen, Yuqing Xia, Chengmin Chi, Guoshuai Zhao, Zheng Liu, Xing Xie, Hao~Allen Sun, Weiwei Deng, Qi~Zhang, and Mao Yang. 2024.
\newblock \href {https://dl.acm.org/doi/10.5555/3600270.3602126} {A neural corpus indexer for document retrieval}.
\newblock In \emph{Proceedings of the 36th International Conference on Neural Information Processing Systems}, NIPS '22, Red Hook, NY, USA. Curran Associates Inc.

\bibitem[{{Xiong} et~al.(2020){Xiong}, {Xiong}, {Li}, {Tang}, {Liu}, {Bennett}, {Ahmed}, and {Overwijk}}]{xiong2020approximate}
Lee {Xiong}, Chenyan {Xiong}, Ye~{Li}, Kwok-Fung {Tang}, Jialin {Liu}, Paul {Bennett}, Junaid {Ahmed}, and Arnold {Overwijk}. 2020.
\newblock \href {https://doi.org/10.48550/arXiv.2007.00808} {{Approximate Nearest Neighbor Negative Contrastive Learning for Dense Text Retrieval}}.
\newblock \emph{arXiv e-prints}, arXiv:2007.00808.

\bibitem[{Xu et~al.(2024)Xu, Sun, Zheng, Geng, Zhao, Feng, Tao, Lin, and Jiang}]{xu2024wizardlm}
Can Xu, Qingfeng Sun, Kai Zheng, Xiubo Geng, Pu~Zhao, Jiazhan Feng, Chongyang Tao, Qingwei Lin, and Daxin Jiang. 2024.
\newblock \href {https://openreview.net/forum?id=CfXh93NDgH} {Wizard{LM}: Empowering large pre-trained language models to follow complex instructions}.
\newblock In \emph{The Twelfth International Conference on Learning Representations}.

\bibitem[{Zhang et~al.(2020)Zhang, Kishore, Wu, Weinberger, and Artzi}]{Zhang2020BERTScore}
Tianyi Zhang, Varsha Kishore, Felix Wu, Kilian~Q. Weinberger, and Yoav Artzi. 2020.
\newblock \href {https://openreview.net/forum?id=SkeHuCVFDr} {Bertscore: Evaluating text generation with bert}.
\newblock In \emph{International Conference on Learning Representations}.

\bibitem[{Zhou et~al.(2023)Zhou, Yao, Dou, Wu, and Wen}]{zhou2023dynamicretriever}
Yu-Jia Zhou, Jing Yao, Zhi-Cheng Dou, Ledell Wu, and Ji-Rong Wen. 2023.
\newblock \href {https://doi.org/10.1007/s11633-022-1373-9} {Dynamicretriever: A pre-trained model-based ir system without an explicit index}.
\newblock \emph{Machine Intelligence Research}, 20(2):276--288.

\bibitem[{Zhuang et~al.(2023)Zhuang, Qin, Jagerman, Hui, Ma, Lu, Ni, Wang, and Bendersky}]{10.1145/3539618.3592047}
Honglei Zhuang, Zhen Qin, Rolf Jagerman, Kai Hui, Ji~Ma, Jing Lu, Jianmo Ni, Xuanhui Wang, and Michael Bendersky. 2023.
\newblock \href {https://doi.org/10.1145/3539618.3592047} {Rankt5: Fine-tuning t5 for text ranking with ranking losses}.
\newblock In \emph{Proceedings of the 46th International ACM SIGIR Conference on Research and Development in Information Retrieval}, SIGIR '23, page 2308–2313, New York, NY, USA. Association for Computing Machinery.

\end{thebibliography}

\clearpage

\appendix

\section{Dataset Details} \label{apx:dataset}
In this section, we present several tables that detail the statistics of the datasets utilized in our study. The tables include comprehensive data such as sample sizes, feature counts, and other relevant metrics, providing an overview of the datasets' composition and scope. 

\begin{table}[ht]
	\center
	\resizebox{\columnwidth}{!}{%
    	\begin{tabular}{@{}lllll@{}}
    		\toprule  Dataset & Scenario & \# of Questions  & \# of Hints  \\ \midrule 
            TriviaQA & Finetuned & 11,313 & 105,709  \\
    		TriviaQA & Vanilla & 11,313 & 103,018  \\ \midrule
            NQ & Finetuned & 3,610 & 33,131  \\
    		NQ & Vanilla & 3,610 & 30,976  \\ \midrule
            WebQ & Finetuned & 2,032 & 16,978  \\
    		WebQ & Vanilla & 2,032 & 15,812  \\
    		\bottomrule
    	\end{tabular}%
     }
	\caption{Statistics of TriviaQA, NQ, and WebQ datasets.}
	\label{tbl:dataset_statistics}
\end{table}

\begin{table}[ht]
	\center
	\begin{tabular}{@{}l|lll@{}}
		\toprule Question Type & TriviaQA & NQ   & WebQ \\
		\midrule Human         & 36\%     & 40\% & 30\% \\
		Location               & 21\%     & 14\% & 28\% \\
		Entity                 & 32\%     & 11\% & 21\% \\
		Description            & 6\%      & 8\%  & 11\% \\
		Other                  & 5\%      & 27\% & 10\% \\
		\bottomrule
	\end{tabular}
	\caption{Distribution of TriviaQA, NQ, and WebQ datasets based on the question type.}
	\label{tbl:dataset_distribution}
\end{table}

\begin{table}[!t]
	\center
	\resizebox{\columnwidth}{!}{%
		\begin{tabular}{@{}p{0.28\textwidth}|lll@{}}
			\toprule
			                             & Training & Validation & Test  \\ 
			\midrule
			Number of questions          & 14,645   & 1,000      & 1,000 \\
			Number of hints              & 140,973  & 9,638      & 9,619 \\
			\midrule
			Avg. question length (words) & 14.18    & 14.08      & 13.95 \\
			Avg. hint length (words)     & 14.98    & 15.07      & 15.14 \\
			Avg. \#hints / question      & 9.62     & 9.63       & 9.61  \\
			Avg. \#entities / question   & 1.35     & 1.40       & 1.35  \\
			Avg. \#entities / hint       & 0.96     & 1.00       & 0.98  \\
			Avg. \#sources / question    & 6.27     & 6.17       & 6.71  \\
			\bottomrule
		\end{tabular}%
	}
	\caption{Statistics of the TriviaHG dataset~\citep{mozafari2024triviahg}}
	\label{tbl:triviahg_statistics}
\end{table}

\section{Metrics}\label{apx:metrics}
In this section, we provide a detailed explanation of the metrics employed in our study to evaluate the effectiveness of our methods. We utilize the scikit-learn library~\citep{scikit-learn} to compute the metrics.

\begin{itemize}
	\item \textbf{Accuracy (ACC):} This metric leverages LLMs to determine the correctness of the answers~\citep{kamalloo-etal-2023-evaluating}.
	\item \textbf{Exact Match (EM):} This metric evaluates whether the retrieved or generated passage perfectly includes the correct answer text without modifications.
	\item \textbf{Precision (PR):} This metric calculates the percentage of words in the retrieved or generated passage that are also found in the correct answer. 
	\item \textbf{Recall (RC):} This metric determines the percentage of words from the correct answer that are included in the retrieved passage. 
	\item \textbf{F1-measure (F1):} This metric is the harmonic mean of precision and recall.
	\item \textbf{Contains (CON):} This metric checks if the retrieved or generated passage encompasses all the correct answer.
	\item \textbf{BERTScore (BERT):} This metric~\citep{Zhang2020BERTScore} calculates the semantic similarity between words in the retrieved passage and the answer, utilizing the contextual embeddings from BERT~\citep{devlin-etal-2019-bert}.
\end{itemize}

\section{Additional Experimental Results} \label{apx:additional_experiments}
In this section, we provide a detailed presentation of the results from our experiments across various scenarios. We will explore how different conditions and variables influence the outcomes. The column \textit{\# of Hints} displays the number of hints used as context, while the column \textit{Ranking} presents various methods for reranking these hints, which are detailed in Section~\ref{ss:hint_generation_methods}.

\begin{table}[!htb]
	\resizebox{\columnwidth}{!}{%
\begin{tabular}{@{}llllllll@{}}
			\toprule \# of Hints & Ranking     & EM    & F1    & PR    & RC    & CON   & BERT  \\
			\midrule \multicolumn{8}{c}{\textit{Zero-Shot}} \\
			\midrule 2           & default     & 20.76 & 25.14 & 26.26 & 25.05 & 26.37 & 75.92 \\
			2                    & convergence & 21.1  & 25.43 & 26.57 & 25.32 & 26.62 & 76.03 \\
			2                    & t5          & 20.22 & 24.6  & 25.73 & 24.48 & 25.77 & 75.78 \\
			5                    & default     & 21.44 & 25.82 & 27.04 & 25.68 & 27.25 & 76.13 \\
			5                    & convergence & 21.44 & 25.72 & 26.97 & 25.54 & 27.15 & 76.03 \\
			5                    & t5          & 20.98 & 25.4  & 26.64 & 25.23 & 26.81 & 76.05 \\
\bottomrule
		\end{tabular}%
	}
	\caption{The results of \textbf{T5-3b} used as the reader, employing zero-shot and few-shot learning strategies on the \textbf{TriviaQA} dataset, are analyzed based on different ranking methods and a range of hint quantities. These hints were generated using the \textbf{HiGen-Va} method.}
	\label{tbl:3b-va-trivia}
\end{table}

\begin{table}[!t]
	\resizebox{\columnwidth}{!}{%
		\begin{tabular}{@{}llllllll@{}}
			\toprule \# of Hints & Ranking     & EM    & F1    & PR    & RC    & CON   & BERT  \\
			\midrule \multicolumn{8}{c}{\textit{Zero-Shot}} \\
			\midrule 2           & default     & 7.92  & 11.03 & 13.18 & 10.4  & 13.32 & 69.81 \\
			2                    & convergence & 8.17  & 11.4  & 13.46 & 10.84 & 13.55 & 69.91 \\
			2                    & t5          & 8.06  & 11.22 & 13.14 & 10.66 & 13.21 & 69.9  \\
			5                    & default     & 8.84  & 12.04 & 14.24 & 11.39 & 14.4  & 70.29 \\
			5                    & convergence & 8.73  & 11.98 & 14.11 & 11.36 & 14.18 & 70.21 \\
			5                    & t5          & 8.12  & 11.46 & 13.52 & 10.87 & 13.6  & 70.11 \\
\bottomrule
		\end{tabular}%
	}
	\caption{The results of \textbf{T5-3b} used as the reader, employing zero-shot and few-shot learning strategies on the \textbf{NQ} dataset, are analyzed based on different ranking methods and a range of hint quantities. These hints were generated using the \textbf{HiGen-Va} method.}
	\label{tbl:3b-va-nq}
\end{table}

\begin{table}[!t]
	\resizebox{\columnwidth}{!}{%
		\begin{tabular}{@{}llllllll@{}}
			\toprule
			\# of Hints & Ranking     & EM    & F1    & PR    & RC    & CON   & BERT  \\ \midrule
			\multicolumn{8}{c}{\textit{Zero-Shot}}                                    \\ \midrule
			2           & default     & 13.24 & 21.09 & 24.99 & 19.65 & 25.15 & 73.1  \\
			2           & convergence & 13.88 & 21.74 & 25.7  & 20.28 & 25.89 & 73.19 \\
			2           & t5          & 13.29 & 21.04 & 25.02 & 19.57 & 24.95 & 73.03 \\
			5           & default     & 13.44 & 21.07 & 25.09 & 19.6  & 25.25 & 73.11 \\
			5           & convergence & 13.09 & 20.75 & 24.82 & 19.23 & 25.05 & 72.96 \\
			5           & t5          & 12.8  & 20.7  & 24.8  & 19.17 & 24.61 & 73.02 \\
\bottomrule
		\end{tabular}%
	}
	\caption{The results of \textbf{T5-3b} used as the reader, employing zero-shot and few-shot learning strategies on the \textbf{WebQ} dataset, are analyzed based on different ranking methods and a range of hint quantities. These hints were generated using the \textbf{HiGen-Va} method.}
	\label{tbl:3b-va-webq}
\end{table}

\begin{table}[!t]
	\resizebox{\columnwidth}{!}{%
		\begin{tabular}{@{}llllllll@{}}
			\toprule \# of Hints & Ranking     & EM    & F1    & PR    & RC    & CON   & BERT  \\
			\midrule \multicolumn{8}{c}{\textit{Zero-Shot}} \\
			\midrule 2           & default     & 21.71 & 26.09 & 27.21 & 25.97 & 27.18 & 76.42 \\
			2                    & convergence & 21.95 & 26.27 & 27.44 & 26.15 & 27.53 & 76.39 \\
			2                    & t5          & 21.4  & 25.79 & 26.94 & 25.67 & 26.83 & 76.3  \\
			5                    & default     & 23.27 & 27.63 & 28.82 & 27.53 & 28.98 & 76.8  \\
			5                    & convergence & 23.13 & 27.46 & 28.68 & 27.33 & 28.89 & 76.79 \\
			5                    & t5          & 22.71 & 27.14 & 28.35 & 27.02 & 28.57 & 76.65 \\
\bottomrule
		\end{tabular}%
	}
	\caption{The results of \textbf{T5-3b} used as the reader, employing zero-shot and few-shot learning strategies on the \textbf{TriviaQA} dataset, are analyzed based on different ranking methods and a range of hint quantities. These hints were generated using the \textbf{HiGen-FT} method.}
	\label{tbl:3b-ft-trivia}
\end{table}

\begin{table}[!t]
	\resizebox{\columnwidth}{!}{%
		\begin{tabular}{@{}llllllll@{}}
			\toprule \# of Hints & Ranking     & EM    & F1    & PR    & RC    & CON   & BERT  \\
			\midrule \multicolumn{8}{c}{\textit{Zero-Shot}} \\
			\midrule 2           & default     & 8.86  & 12.4  & 14.4  & 11.82 & 14.18 & 70.8  \\
			2                    & convergence & 9.28  & 12.69 & 14.74 & 12.09 & 14.65 & 70.96 \\
			2                    & t5          & 8.81  & 12.35 & 14.46 & 11.72 & 14.35 & 70.93 \\
			5                    & default     & 10.19 & 13.97 & 16.18 & 13.33 & 15.87 & 71.68 \\
			5                    & convergence & 10.44 & 14.32 & 16.59 & 13.67 & 16.4  & 71.72 \\
			5                    & t5          & 10.19 & 13.88 & 16.21 & 13.18 & 16.04 & 71.64 \\
\bottomrule
		\end{tabular}%
	}
	\caption{The results of \textbf{T5-3b} used as the reader, employing zero-shot and few-shot learning strategies on the \textbf{NQ} dataset, are analyzed based on different ranking methods and a range of hint quantities. These hints were generated using the \textbf{HiGen-FT} method.}
	\label{tbl:3b-ft-nq}
\end{table}

\begin{table}[!t]
	\resizebox{\columnwidth}{!}{%
		\begin{tabular}{@{}llllllll@{}}
			\toprule \# of Hints & Ranking     & EM    & F1    & PR    & RC    & CON   & BERT  \\
			\midrule \multicolumn{8}{c}{\textit{Zero-Shot}} \\
			\midrule 2           & default     & 13.19 & 21.16 & 25.3  & 19.6  & 25.94 & 72.9  \\
			2                    & convergence & 13.58 & 21.82 & 26.09 & 20.2  & 26.38 & 73.07 \\
			2                    & t5          & 13.24 & 21.63 & 25.81 & 20.11 & 26.18 & 73.31 \\
			5                    & default     & 13.88 & 22.07 & 26.38 & 20.4  & 27.17 & 73.56 \\
			5                    & convergence & 14.12 & 22.23 & 26.45 & 20.65 & 26.82 & 73.53 \\
			5                    & t5          & 13.78 & 22.07 & 26.36 & 20.44 & 26.57 & 73.42 \\
\bottomrule
		\end{tabular}%
	}
	\caption{The results of \textbf{T5-3b} used as the reader, employing zero-shot and few-shot learning strategies on the \textbf{WebQ} dataset, are analyzed based on different ranking methods and a range of hint quantities. These hints were generated using the \textbf{HiGen-FT} method.}
	\label{tbl:3b-ft-webq}
\end{table}

\begin{table}[!t]
	\resizebox{\columnwidth}{!}{%
		\begin{tabular}{@{}lllllllll@{}}
			\toprule \# of Hints & Ranking     & ACC   & EM    & F1    & PR    & RC    & CON   & BERT  \\
			\midrule \multicolumn{9}{c}{\textit{Zero-Shot}} \\
			\midrule 2           & default     & 56.12 & 0     & 7.99  & 4.69  & 43.36 & 53.48 & 70.25 \\
			2                    & convergence & 55.98 & 0     & 8     & 4.7   & 43.49 & 54.11 & 70.26 \\
			2                    & t5          & 54.16 & 0     & 7.93  & 4.67  & 41.85 & 51.55 & 70.08 \\
			5                    & default     & 58.25 & 0     & 7.91  & 4.66  & 41.87 & 54.48 & 70.28 \\
			5                    & convergence & 58.25 & 0     & 7.93  & 4.67  & 42.34 & 54.76 & 70.26 \\
			5                    & t5          & 57.56 & 0     & 7.85  & 4.64  & 41.33 & 53    & 70.12 \\
\bottomrule
		\end{tabular}%
	}
	\caption{The results of \textbf{LLaMA-7b} used as the reader, employing zero-shot and few-shot learning strategies on the \textbf{TriviaQA} dataset, are analyzed based on different ranking methods and a range of hint quantities. These hints were generated using the \textbf{HiGen-Va} method.}
	\label{tbl:llama-va-trivia}
\end{table}

\begin{table}[!t]
	\resizebox{\columnwidth}{!}{%
		\begin{tabular}{@{}lllllllll@{}}
			\toprule \# of Hints & Ranking     & ACC   & EM    & F1    & PR    & RC    & CON   & BERT  \\
			\midrule \multicolumn{9}{c}{\textit{Zero-Shot}} \\
			\midrule 2           & default     & 48.25 & 0     & 3.16  & 1.77  & 25.37 & 23.52 & 64.47 \\
			2                    & convergence & 50.55 & 0     & 3.29  & 1.84  & 25.97 & 24.88 & 64.56 \\
			2                    & t5          & 47.56 & 0     & 3.18  & 1.78  & 24.15 & 22.08 & 64.34 \\
			5                    & default     & 48.67 & 0     & 3.35  & 1.89  & 25.22 & 25.18 & 64.86 \\
			5                    & convergence & 50.36 & 0     & 3.48  & 1.97  & 25.74 & 26.12 & 65    \\
			5                    & t5          & 48.75 & 0     & 3.3   & 1.87  & 25.11 & 24.43 & 64.68 \\
\bottomrule
		\end{tabular}%
	}
	\caption{The results of \textbf{LLaMA-7b} used as the reader, employing zero-shot and few-shot learning strategies on the \textbf{NQ} dataset, are analyzed based on different ranking methods and a range of hint quantities. These hints were generated using the \textbf{HiGen-Va} method.}
	\label{tbl:llama-va-nq}
\end{table}

\begin{table}[!t]
	\resizebox{\columnwidth}{!}{%
		\begin{tabular}{@{}lllllllll@{}}
			\toprule \# of Hints & Ranking     & ACC   & EM    & F1    & PR    & RC    & CON   & BERT  \\
			\midrule \multicolumn{9}{c}{\textit{Zero-Shot}} \\
			\midrule 2           & default     & 51.62 & 0     & 5.51  & 3.23  & 36.43 & 35.73 & 66.98 \\
			2                    & convergence & 53.49 & 0     & 5.4   & 3.17  & 35.32 & 35.93 & 66.9  \\
			2                    & t5          & 52.36 & 0     & 5.41  & 3.18  & 35.35 & 33.96 & 66.75 \\
			5                    & default     & 50.94 & 0     & 5.75  & 3.38  & 36.99 & 39.07 & 67.5  \\
			5                    & convergence & 51.43 & 0     & 5.73  & 3.36  & 37.33 & 39.27 & 67.43 \\
			5                    & t5          & 50.15 & 0     & 5.65  & 3.32  & 35.91 & 36.91 & 67.28 \\
\bottomrule
		\end{tabular}%
	}
	\caption{The results of \textbf{LLaMA-7b} used as the reader, employing zero-shot and few-shot learning strategies on the \textbf{WebQ} dataset, are analyzed based on different ranking methods and a range of hint quantities. These hints were generated using the \textbf{HiGen-Va} method.}
	\label{tbl:llama-va-webq}
\end{table}

\begin{table}[!t]
	\resizebox{\columnwidth}{!}{%
		\begin{tabular}{@{}lllllllll@{}}
			\toprule \# of Hints & Ranking     & ACC   & EM    & F1    & PR    & RC    & CON   & BERT  \\
			\midrule \multicolumn{9}{c}{\textit{Zero-Shot}} \\
			\midrule 2           & default     & 49.89 & 0     & 8.85  & 5.23  & 44.22 & 59.71 & 71.18 \\
			2                    & convergence & 49.34 & 0     & 8.69  & 5.14  & 44.1  & 59.19 & 71.02 \\
			2                    & t5          & 50.69 & 0     & 8.68  & 5.13  & 43.23 & 56.85 & 70.93 \\
			5                    & default     & 53.58 & 0     & 8.86  & 5.26  & 43.17 & 61.02 & 71.33 \\
			5                    & convergence & 53.91 & 0     & 8.74  & 5.18  & 43.48 & 60.74 & 71.21 \\
			5                    & t5          & 54    & 0     & 8.69  & 5.15  & 42.95 & 59.49 & 71.14 \\
\bottomrule
		\end{tabular}%
	}
	\caption{The results of \textbf{LLaMA-7b} used as the reader, employing zero-shot and few-shot learning strategies on the \textbf{TriviaQA} dataset, are analyzed based on different ranking methods and a range of hint quantities. These hints were generated using the \textbf{HiGen-FT} method.}
	\label{tbl:llama-ft-trivia}
\end{table}

\begin{table}[!t]
	\resizebox{\columnwidth}{!}{%
		\begin{tabular}{@{}lllllllll@{}}
			\toprule \# of Hints & Ranking     & ACC   & EM    & F1    & PR    & RC    & CON   & BERT  \\
			\midrule \multicolumn{9}{c}{\textit{Zero-Shot}} \\
			\midrule 2           & default     & 39.62 & 0     & 3.87  & 2.19  & 26.58 & 29.86 & 66.12 \\
			2                    & convergence & 38.62 & 0     & 3.82  & 2.16  & 26.27 & 30.36 & 65.95 \\
			2                    & t5          & 42.14 & 0     & 3.86  & 2.18  & 25.74 & 27.48 & 65.82 \\
			5                    & default     & 47.32 & 0     & 4.32  & 2.45  & 27.45 & 32.94 & 66.7  \\
			5                    & convergence & 46.65 & 0     & 4.27  & 2.43  & 27.77 & 32.91 & 66.55 \\
			5                    & t5          & 47.76 & 0     & 4.11  & 2.34  & 26.34 & 30.94 & 66.36 \\
\bottomrule
		\end{tabular}%
	}
	\caption{The results of \textbf{LLaMA-7b} used as the reader, employing zero-shot and few-shot learning strategies on the \textbf{NQ} dataset, are analyzed based on different ranking methods and a range of hint quantities. These hints were generated using the \textbf{HiGen-FT} method.}
	\label{tbl:llama-ft-nq}
\end{table}

\begin{table}[!t]
	\resizebox{\columnwidth}{!}{%
		\begin{tabular}{@{}lllllllll@{}}
			\toprule \# of Hints & Ranking     & ACC   & EM    & F1    & PR    & RC    & CON   & BERT  \\
			\midrule \multicolumn{9}{c}{\textit{Zero-Shot}} \\
			\midrule 2           & default     & 47.33 & 0     & 6.23  & 3.65  & 38.76 & 41.49 & 67.96 \\
			2                    & convergence & 46.25 & 0     & 6.19  & 3.64  & 39.29 & 41.54 & 67.81 \\
			2                    & t5          & 49.75 & 0     & 6.2   & 3.62  & 37.08 & 38.48 & 67.91 \\
			5                    & default     & 52.7  & 0     & 6.7   & 3.93  & 39.31 & 43.7  & 68.58 \\
			5                    & convergence & 52.45 & 0     & 6.66  & 3.92  & 39.18 & 43.21 & 68.48 \\
			5                    & t5          & 52.6  & 0     & 6.68  & 3.94  & 38.38 & 42.18 & 68.48 \\
\bottomrule
		\end{tabular}%
	}
	\caption{The results of \textbf{LLaMA-7b} used as the reader, employing zero-shot and few-shot learning strategies on the \textbf{WebQ} dataset, are analyzed based on different ranking methods and a range of hint quantities. These hints were generated using the \textbf{HiGen-FT} method.}
	\label{tbl:llama-ft-webq} 
\end{table}

\begin{table}[!t]
	\resizebox{\columnwidth}{!}{%
		\begin{tabular}{@{}lllllllll@{}}
			\toprule
			\# of Hints & Ranking     & ACC & EM & F1    & PR    & RC    & CON & BERT  \\ \midrule
			\multicolumn{9}{c}{\textit{Zero-Shot}}                                     \\ \midrule
			2           & default     & 78  & 0  & 11.26 & 6.62  & 54.65 & 73  & 71.7  \\
			2           & convergence & 78  & 0  & 11.26 & 6.62  & 54.65 & 73  & 71.71 \\
			2           & t5          & 79  & 0  & 11.39 & 6.75  & 54.58 & 74  & 71.87 \\
			5           & default     & 76  & 0  & 10.85 & 6.52  & 47.73 & 74  & 71.71 \\
			5           & convergence & 76  & 0  & 10.85 & 6.52  & 47.73 & 74  & 71.72 \\
			5           & t5          & 73  & 0  & 10.62 & 6.44  & 44.65 & 67  & 71.45 \\
\bottomrule
		\end{tabular}%
	}
	\caption{The results of \textbf{LLaMA-7b} used as the reader, employing zero-shot and few-shot learning strategies on 100 random questions, are analyzed based on different ranking methods and a range of hint quantities. These hints were generated using the \textbf{LLaMA-7b-FT} core.}
	\label{tbl:ablation_study_llama_7b_ft}
\end{table}

\begin{table}[!t]
	\resizebox{\columnwidth}{!}{%
		\begin{tabular}{@{}lllllllll@{}}
			\toprule
			\multicolumn{1}{c}{\# of Hints} & Ranking     & ACC & EM & F1    & PR    & RC    & CON & BERT  \\ \midrule
			\multicolumn{9}{c}{\textit{Zero-Shot}}                                                         \\ \midrule
			2                               & default     & 68  & 0  & 9.37  & 5.57  & 47.48 & 65  & 70.51 \\
			2                               & convergence & 68  & 0  & 9.37  & 5.57  & 47.48 & 65  & 70.52 \\
			2                               & t5          & 70  & 0  & 9.55  & 5.68  & 46.37 & 60  & 70.38 \\
			5                               & default     & 67  & 0  & 9.15  & 5.43  & 44.4  & 60  & 70.6  \\
			5                               & convergence & 67  & 0  & 9.15  & 5.43  & 44.4  & 60  & 70.62 \\
			5                               & t5          & 69  & 0  & 9.52  & 5.63  & 45.65 & 59  & 70.74 \\
\bottomrule
		\end{tabular}%
	}
	\caption{The results of \textbf{LLaMA-7b} used as the reader, employing zero-shot and few-shot learning strategies on 100 random questions, are analyzed based on different ranking methods and a range of hint quantities. These hints were generated using the \textbf{LLaMA-7b-Va} core.}
	\label{tbl:ablation_study_llama_7b_va}
\end{table}

\begin{table}[!t]
	\resizebox{\columnwidth}{!}{%
		\begin{tabular}{@{}lllllllll@{}}
			\toprule
			\# of Hints & Ranking     & ACC & EM & F1    & PR    & RC    & CON & BERT  \\ \midrule
			\multicolumn{9}{c}{\textit{Zero-Shot}}                                     \\ \midrule
			2           & default     & 77  & 0  & 9.79  & 5.67  & 52.41 & 71  & 71.2  \\
			2           & convergence & 77  & 0  & 9.79  & 5.67  & 52.41 & 71  & 71.2  \\
			2           & t5          & 79  & 0  & 9.98  & 5.91  & 48.09 & 73  & 71.39 \\
			5           & default     & 77  & 0  & 9.77  & 5.76  & 46.86 & 69  & 71.35 \\
			5           & convergence & 77  & 0  & 9.86  & 5.81  & 46.86 & 69  & 71.42 \\
			5           & t5          & 78  & 0  & 9.64  & 5.67  & 47.24 & 70  & 71.41 \\
\bottomrule
		\end{tabular}%
	}
	\caption{The results of \textbf{LLaMA-7b} used as the reader, employing zero-shot and few-shot learning strategies on 100 random questions, are analyzed based on different ranking methods and a range of hint quantities. These hints were generated using the \textbf{LLaMA-13b-Va} core.}
	\label{tbl:ablation_study_llama_13b_va}
\end{table}

\begin{table}[!t]
	\resizebox{\columnwidth}{!}{%
		\begin{tabular}{@{}lllllllll@{}}
			\toprule
			\# of Hints & Ranking     & ACC & EM & F1    & PR    & RC    & CON & BERT  \\ \midrule
			\multicolumn{9}{c}{\textit{Zero-Shot}}                                     \\ \midrule
			2           & default     & 76  & 0  & 10.61 & 6.29  & 49.1  & 71  & 71.72 \\
			2           & convergence & 76  & 0  & 10.71 & 6.35  & 49.02 & 71  & 71.74 \\
			2           & t5          & 75  & 0  & 10.27 & 6.09  & 49.53 & 68  & 71.47 \\
			5           & default     & 83  & 0  & 10.66 & 6.31  & 48.23 & 75  & 71.68 \\
			5           & convergence & 83  & 0  & 10.65 & 6.31  & 48.23 & 76  & 71.68 \\
			5           & t5          & 78  & 0  & 10.28 & 6.1   & 49.52 & 70  & 71.47 \\
\bottomrule
		\end{tabular}%
	}
	\caption{The results of \textbf{LLaMA-7b} used as the reader, employing zero-shot and few-shot learning strategies on 100 random questions, are analyzed based on different ranking methods and a range of hint quantities. These hints were generated using the \textbf{LLaMA-13b-FT} core.}
	\label{tbl:ablation_study_llama_13b_ft}
\end{table}

\begin{table}[!t]
	\resizebox{\columnwidth}{!}{%
		\begin{tabular}{@{}lllllllll@{}}
			\toprule
			\# of Hints & Ranking     & ACC & EM & F1    & PR    & RC    & CON & BERT  \\ \midrule
			\multicolumn{9}{c}{\textit{Zero-Shot}}                                     \\ \midrule
			2           & default     & 78  & 0  & 10.12 & 5.84  & 53.18 & 77  & 71.44 \\
			2           & convergence & 78  & 0  & 10.12 & 5.84  & 53.18 & 77  & 71.44 \\
			2           & t5          & 78  & 0  & 9.98  & 5.91  & 47.53 & 72  & 71.39 \\
			5           & default     & 81  & 0  & 10.49 & 6.25  & 45.47 & 77  & 71.95 \\
			5           & convergence & 81  & 0  & 10.49 & 6.25  & 46.13 & 77  & 71.95 \\
			5           & t5          & 81  & 0  & 10.71 & 6.42  & 45.52 & 72  & 71.52 \\
\bottomrule
		\end{tabular}%
	}
	\caption{The results of \textbf{LLaMA-7b} used as the reader, employing zero-shot and few-shot learning strategies on 100 random questions, are analyzed based on different ranking methods and a range of hint quantities. These hints were generated using the \textbf{LLaMA-70b-Va} core.}
	\label{tbl:ablation_study_llama_70b_va}
\end{table}

\begin{table}[!t]
	\resizebox{\columnwidth}{!}{%
		\begin{tabular}{@{}lllllllll@{}}
			\toprule
			\# of Hints & Ranking     & ACC & EM & F1    & PR    & RC    & CON & BERT  \\ \midrule
			\multicolumn{9}{c}{\textit{Zero-Shot}}                                     \\ \midrule
			2           & default     & 78  & 0  & 10.54 & 6.2   & 50.61 & 70  & 71.43 \\
			2           & convergence & 78  & 0  & 10.55 & 6.2   & 50.61 & 70  & 71.45 \\
			2           & t5          & 75  & 0  & 10.69 & 6.33  & 48.28 & 73  & 71.6  \\
			5           & default     & 79  & 0  & 10.11 & 6.09  & 43.37 & 73  & 71.73 \\
			5           & convergence & 79  & 0  & 10.07 & 6.06  & 43.57 & 73  & 71.74 \\
			5           & t5          & 80  & 0  & 9.94  & 5.9   & 47.22 & 75  & 71.58 \\
\bottomrule
		\end{tabular}%
	}
	\caption{The results of \textbf{LLaMA-7b} used as the reader, employing zero-shot and few-shot learning strategies on 100 random questions, are analyzed based on different ranking methods and a range of hint quantities. These hints were generated using the \textbf{WizardLM-70b} core.}
	\label{tbl:ablation_study_winzardlm_70b}
\end{table}

\begin{table}[!t]
	\resizebox{\columnwidth}{!}{%
		\begin{tabular}{@{}lllllllll@{}}
			\toprule
			\# of Hints & Ranking     & ACC & EM & F1    & PR    & RC    & CON & BERT  \\ \midrule
			\multicolumn{9}{c}{\textit{Zero-Shot}}                                     \\ \midrule
			2           & default     & 81  & 0  & 10.6  & 6.13  & 58.87 & 81  & 71.45 \\
			2           & convergence & 81  & 0  & 10.6  & 6.13  & 59.37 & 81  & 71.45 \\
			2           & t5          & 79  & 0  & 9.94  & 5.81  & 53.97 & 77  & 71.55 \\
			5           & default     & 83  & 0  & 10.25 & 6.07  & 50.38 & 77  & 71.55 \\
			5           & convergence & 83  & 0  & 10.25 & 6.07  & 50.38 & 78  & 71.59 \\
			5           & t5          & 82  & 0  & 10.17 & 6.03  & 46.72 & 79  & 72.14 \\
\bottomrule
		\end{tabular}%
	}
	\caption{The results of \textbf{LLaMA-7b} used as the reader, employing zero-shot and few-shot learning strategies on 100 random questions, are analyzed based on different ranking methods and a range of hint quantities. These hints were generated using the \textbf{GPT 3.5} core.}
	\label{tbl:ablation_study_gpt3.5}
\end{table}

\begin{table}[!t]
	\resizebox{\columnwidth}{!}{%
		\begin{tabular}{@{}lllllllll@{}}
			\toprule
			\# of Hints & Ranking     & ACC & EM & F1    & PR    & RC    & CON & BERT  \\ \midrule
			\multicolumn{9}{c}{\textit{Zero-Shot}}                                     \\ \midrule
			2           & default     & 84  & 0  & 11.07 & 6.55  & 53.63 & 80  & 72.23 \\
			2           & convergence & 84  & 0  & 11.06 & 6.55  & 53.13 & 79  & 72.22 \\
			2           & t5          & 81  & 0  & 11.59 & 6.88  & 52.93 & 79  & 72.39 \\
			5           & default     & 86  & 0  & 11.5  & 6.82  & 53.07 & 82  & 72.54 \\
			5           & convergence & 86  & 0  & 11.48 & 6.81  & 53.07 & 82  & 72.56 \\
			5           & t5          & 83  & 0  & 11.81 & 7.02  & 51.35 & 79  & 72.36 \\
\bottomrule
		\end{tabular}%
	}
	\caption{The results of \textbf{LLaMA-7b} used as the reader, employing zero-shot and few-shot learning strategies on 100 random questions, are analyzed based on different ranking methods and a range of hint quantities. These hints were generated using the \textbf{Gemini} core.}
	\label{tbl:ablation_study_gemini}
\end{table}

\begin{table}[!t]
	\resizebox{\columnwidth}{!}{%
		\begin{tabular}{@{}lllllllll@{}}
			\toprule
			\# of Hints & Ranking     & ACC & EM & F1    & PR    & RC    & CON & BERT  \\ \midrule
			\multicolumn{9}{c}{\textit{Zero-Shot}}                                     \\ \midrule
			2           & default     & 83  & 0  & 11.27 & 6.76  & 50.72 & 81  & 72.39 \\
			2           & convergence & 83  & 0  & 11.28 & 6.77  & 50.72 & 81  & 72.41 \\
			2           & t5          & 80  & 0  & 10.54 & 6.26  & 48.2  & 71  & 71.83 \\
			5           & default     & 82  & 0  & 11.3  & 6.76  & 50.85 & 79  & 72.42 \\
			5           & convergence & 82  & 0  & 11.3  & 6.76  & 50.85 & 79  & 72.42 \\
			5           & t5          & 85  & 0  & 11.71 & 7.03  & 49.68 & 76  & 72.32 \\
\bottomrule
		\end{tabular}%
	}
	\caption{The results of \textbf{LLaMA-7b} used as the reader, employing zero-shot and few-shot learning strategies on 100 random questions, are analyzed based on different ranking methods and a range of hint quantities. These hints were generated using the \textbf{LLaMA-70b-FT} core.}
	\label{tbl:ablation_study_llama_70b_ft}
\end{table}

\begin{table}[!t]
	\resizebox{\columnwidth}{!}{%
		\begin{tabular}{@{}lllllllll@{}}
			\toprule
			\# of Hints & Ranking     & ACC & EM & F1    & PR    & RC    & CON & BERT  \\ \midrule
			\multicolumn{9}{c}{\textit{Zero-Shot}}                                     \\ \midrule
			2           & default     & 87  & 0  & 11.91 & 7.07  & 56.4  & 82  & 72.3  \\
			2           & convergence & 87  & 0  & 11.89 & 7.06  & 56.4  & 82  & 72.28 \\
			2           & t5          & 89  & 0  & 11.67 & 7.02  & 58.62 & 83  & 72.1  \\
			5           & default     & 92  & 0  & 11.58 & 6.9   & 50.98 & 83  & 72.75 \\
			5           & convergence & 92  & 0  & 11.58 & 6.9   & 50.98 & 83  & 72.75 \\
			5           & t5          & 90  & 0  & 11.65 & 6.99  & 55.4  & 84  & 72.54 \\
\bottomrule
		\end{tabular}%
	}
	\caption{The results of \textbf{LLaMA-7b} used as the reader, employing zero-shot and few-shot learning strategies on 100 random questions, are analyzed based on different ranking methods and a range of hint quantities. These hints were generated using the \textbf{Copilot} core.}
	\label{tbl:ablation_study_copilot}
\end{table}

\begin{table}[!t]
	\resizebox{\columnwidth}{!}{%
		\begin{tabular}{@{}lllllllll@{}}
			\toprule
			\# of Hints & Ranking     & ACC & EM & F1    & PR    & RC    & CON & BERT  \\ \midrule
			\multicolumn{9}{c}{\textit{Zero-Shot}}                                     \\ \midrule
			2           & default     & 85  & 0  & 10.38 & 6.11  & 53.5  & 81  & 71.85 \\
			2           & convergence & 85  & 0  & 10.38 & 6.11  & 53.5  & 81  & 71.85 \\
			2           & t5          & 85  & 0  & 10.83 & 6.34  & 53.9  & 80  & 72.11 \\
			5           & default     & 96  & 0  & 11.5  & 6.8   & 53.97 & 89  & 73.2  \\
			5           & convergence & 96  & 0  & 11.5  & 6.8   & 53.97 & 89  & 73.2  \\
			5           & t5          & 88  & 0  & 11.78 & 6.92  & 52.62 & 81  & 72.58 \\
\bottomrule
		\end{tabular}%
	}
	\caption{The results of \textbf{LLaMA-7b} used as the reader, employing zero-shot and few-shot learning strategies on 100 random questions, are analyzed based on different ranking methods and a range of hint quantities. These hints were generated using the \textbf{GPT 4} core.}
	\label{tbl:ablation_study_gpt_4}
\end{table}

\begin{table}[!t]
	\resizebox{\columnwidth}{!}{%
		\begin{tabular}{@{}l|lllll@{}}
			\toprule Hint Generator & \# of Parameters & \# of Hints & Ranking & EM            & F1             \\
			\midrule \multicolumn{6}{c}{\textit{Zero-Shot}} \\
			\midrule LLaMA-Va       & 7b               & 10          & default & 33.0          & 36.82          \\
			WizardLM 70b            & 70b              & 7           & t5      & 34.0          & 37.02          \\
			LLaMA-FT                & 7b               & 2           & default & 35.0          & 39.43          \\
			LLaMA-Va                & 13b              & 7           & default & 35.0          & 40.39          \\
			LLaMA-Va                & 70b              & 10          & default & 35.0          & 37.63          \\
			LLaMA-FT                & 13b              & 2           & t5      & 38.0          & 40.73          \\
\bottomrule
		\end{tabular}%
	}
	\caption{The performance of T5-3b across different LLMs as the central component of the \method system, generating hints for 100 questions.}
	\label{tbl:ablation_study_3b}
\end{table}

\cleardoublepage
\section{Case Studies} \label{apx:case_study}
In this section, we show several case studies that illustrate the prompts we have chosen, along with examples from our experiments and their respective outcomes. The case studies are designed to demonstrate the practical application of our theoretical framework and to showcase the effectiveness of our chosen methodologies in real-world scenarios.

\begin{table*}
	\resizebox{\textwidth}{!}{%
		\begin{tabular}{@{}p{5cm}|lll|l@{}}
			\toprule
			Question                                                                                                 & Retriever                    & LLaMA-70b             & \method                      & True Answer                  \\ \midrule
			\multicolumn{5}{c}{\textit{TriviaQA}}                                                                                                                                                                                         \\ \midrule
			How many dot positions are usually used in each letter of the Braille system?                            & \textbf{6}                   & \textbf{six}          & \textbf{six}                 & \textbf{6, six}              \\ \midrule
			Who was the leader of the gang whose members included Benny the Ball ,Brain and Choo Choo?               & the bowery boys              & \textbf{top cat}      & \textbf{top cat}             & \textbf{top cat}             \\ \midrule
			Which Glasgow group signed to Creation Records and recorded their debut single "All Fall Down", in 1985? & \textbf{primal scream}       & the pastels           & the jesus and mary chain     & \textbf{primal scream}       \\ \midrule
			Who is the only man to win a best actor Oscar playing brothers?                                          & jack nicholson               & daniel day            & henry fonda                  & \textbf{lee marvin}          \\ \midrule
			\multicolumn{5}{c}{\textit{NQ}}                                                                                                                                                                                               \\ \midrule
			who played taylor on the bold and beautiful?                                                             & \textbf{hunter tylo}         & \textbf{hunter tylo}  & \textbf{hunter tylo}         & \textbf{hunter tylo}         \\ \midrule
			who wrote the song going to kansas city?                                                                 & bo diddley                   & \textbf{jerry leiber} & \textbf{jerry leiber}        & \textbf{jerry leiber}        \\ \midrule
			what part of the brain is in the middle cranial fossa?                                                   & \textbf{the pituitary gland} & temporal lobe region  & the hippocampus              & \textbf{the pituitary gland} \\ \midrule
			who did the broncos beat in the super bowl?                                                              & the packers                  & green bay             & the falcons                  & \textbf{carolina panthers}   \\ \midrule
			\multicolumn{5}{c}{\textit{WebQ}}                                                                                                                                                                                             \\ \midrule
			where are boeing headquarters?                                                                           & \textbf{chicago}             & \textbf{chicago}      & \textbf{seattle}             & \textbf{seattle, chicago}    \\ \midrule
			what university did obama graduated from?                                                                & harvard law school           & harvard law school    & \textbf{columbia university} & \textbf{columbia university} \\ \midrule
			what country did buddha come from?                                                                       & \textbf{india}               & india                 & nepal                        & \textbf{india}               \\ \midrule
			who played amy squirrel in bad teacher?                                                                  & cameron diaz                 & \textbf{lucy punch}   & cameron diaz                 & \textbf{lucy punch}          \\ \bottomrule
		\end{tabular}%
	}
	\caption{Comparison of answers for randomly selected questions from the TriviaQA, NQ, and WebQ datasets.}
	\label{tbl:case_study_trivia_nq_webq}
\end{table*}

\begin{table*}[p]
	\resizebox{\textwidth}{!}{%
		\begin{tabular}{@{}ll@{}}
			\toprule
			\multicolumn{2}{l}{\textbf{Question:} what city of USA has a neighborhood called little havana?}                                                                                                                                          \\ \midrule
			\multicolumn{2}{l}{\textbf{Answer:} Miami}                                                                                                                                                                                         \\ \\
			\multicolumn{2}{l}{\begin{tabular}[c]{@{}l@{}}\textbf{Candidate Answers:} \\ 1. Havana\\ 2. Washington D.C.\\ 3. San Francisco\\ 4. Chicago\\ 5. New York\\ 6. Los Angeles\\ 7. Houston\\ 8. Dallas\\ 9. Atlanta\\ 10. Philadelphia\\ 11. Miami\end{tabular}} \\ \midrule
			\multicolumn{2}{c}{\textbf{Vanilla}}                                                                                                                                                                                               \\ \midrule
			\multicolumn{1}{l|}{\textbf{Hints}}                                                                    & \textbf{Candidates}   \\ \midrule
			\multicolumn{1}{l|}{This city is located in the southeastern United States.}                           & 0 0 0 0 0 0 0 0 1 1 1 \\
			\multicolumn{1}{l|}{The city is known for its vibrant Cuban culture and cuisine.}                      & 1 0 0 0 0 0 0 0 0 0 0 \\
			\multicolumn{1}{l|}{The city has a large Hispanic population.}                                         & 0 1 0 1 1 1 1 1 1 1 1 \\
			\multicolumn{1}{l|}{The neighborhood is home to many Cuban restaurants and shops.}                     & 0 0 0 0 0 0 0 0 0 0 0 \\
			\multicolumn{1}{l|}{The city is located near the Atlantic Ocean.}                                      & 0 1 0 0 1 0 0 0 1 1 1 \\
			\multicolumn{1}{l|}{The city has a tropical climate.}                                                  & 1 0 0 0 0 0 0 0 0 0 1 \\
			\multicolumn{1}{l|}{The neighborhood is known for its colorful street art and murals.}                 & 0 1 1 0 0 1 0 0 0 1 1 \\
			\multicolumn{1}{l|}{The city has a rich history of Cuban immigration.}                                 & 0 0 0 0 1 0 0 0 0 0 1 \\
			\multicolumn{1}{l|}{The neighborhood is a popular destination for foodies and cultural enthusiasts.}   & 0 0 1 0 1 0 0 0 0 0 1 \\ \midrule
			\multicolumn{2}{c}{\textbf{Finetuned}}                                                                                                                                                                                             \\ \midrule
			\multicolumn{1}{l|}{\textbf{Hints}}                                                                    & \textbf{Candidates}   \\ \midrule
			\multicolumn{1}{l|}{It is the largest city in Florida.}                                                & 0 0 0 0 0 0 0 0 0 0 0 \\
			\multicolumn{1}{l|}{It’s known for its vibrant nightlife, especially in a place called South Beach.} & 0 0 0 0 0 0 0 0 0 0 1 \\
			\multicolumn{1}{l|}{It’s the only major U.S. city to be founded by a woman.}                         & 0 0 0 0 0 0 0 0 0 0 1 \\
			\multicolumn{1}{l|}{It’s home to one of the largest cruise ship ports in the world.}                 & 0 0 0 0 1 1 1 0 0 0 1 \\
			\multicolumn{1}{l|}{It is nicknamed the "Capital of Latin America".}                                   & 1 0 0 0 0 0 0 0 0 0 1 \\
			\multicolumn{1}{l|}{The city is known for its Art Deco Historic District.}                             & 1 0 0 0 0 0 0 0 0 0 1 \\
			\multicolumn{1}{l|}{The city is often at risk from hurricanes due to its location.}                    & 1 0 0 0 1 0 1 0 0 0 1 \\
			\multicolumn{1}{l|}{It is located in the southeastern part of the state.}                              & 0 0 0 0 0 0 0 0 1 0 0 \\
			\multicolumn{1}{l|}{It is the 44th-most populous city in the United States.}                           & 0 0 0 1 1 0 1 0 1 1 0 \\
			\multicolumn{1}{l|}{It is the 16th-most populous metropolitan area in the United States.}              & 0 0 0 1 1 1 1 1 1 1 0 \\ \bottomrule
		\end{tabular}%
	}
	\caption{Hints generated by the HiGen-Va and HiGen-FT methods and the candidate answers they encompass. The 'Candidates' column displays which candidate answers are included in each hint, indicated by the index of candidate answers; for example, the first bit confirms the inclusion of 'Havana' and so on.}
	\label{tbl:case_study_candidates}
\end{table*}

\begin{table*}[p]
\begin{tabular}{@{}p{\textwidth}@{}}
		\toprule                                                                                                                                                                                                                                                                                                                                                                                                                                                                                                                                                                                                                                                                                                                                 
		\textbf{Question:} what city of USA has a neighborhood called little havana?                                                                                                                                                                                                                                                                                                                                                                                                                                                                                                                                                                                                                                                             \\
\bottomrule                                                                                                                                                                                                                                                                                                                                                                                                                                                                                                                                                                                                                                                                                                                              
	\end{tabular}
	\caption{Case study of the retrieved passage from MSS-DPR, generated context by LLaMA-70b, and hints generated by \method on LLaMA 7b in Zero-Shot. Words in \textcolor{blue}{blue} indicate the correct answer, while those in \textcolor{red}{red} represent other potential answers.}
	\label{tbl:case_study_retriever_genrative_higenqa_zero}
\end{table*}

\begin{table*}[p]
	\small
	\begin{tabular}{@{}p{\textwidth}@{}}
		\toprule                                                                                                                                                                                                                                                                                                                                                                                                                                                                                                                                                                                                                                                                                                                                                                    
		\textbf{Question:} what city of USA has a neighborhood called little havana?                                                                                                                                                                                                                                                                                                                                                                                                                                                                                                                                                                                                                                                                                                \\
\bottomrule                                                                                                                                                                                                                                                                                                                                                                                                                                                                                                                                                                                                                                                                                                                                                                 
	\end{tabular}
	\caption{Case study of the retrieved passage from MSS-DPR retriever on LLaMA-7b in Few-Shot. Words in \textcolor{blue}{blue} indicate the correct answer, while those in \textcolor{red}{red} represent other potential answers.}
	\label{tbl:case_study_retriever}
\end{table*}

\begin{table*}[p]
	\small
	\begin{tabular}{@{}p{\textwidth}@{}}
		\toprule                                                                                                                                                                                                                                                                                                                                                                                                                                                                                                                                                                                                                                                                                                                                                                                                                                                                                                  
		\textbf{Question:} what city of USA has a neighborhood called little havana?                                                                                                                                                                                                                                                                                                                                                                                                                                                                                                                                                                                                                                                                                                                                                                                                                              \\
\bottomrule                                                                                                                                                                                                                                                                                                                                                                                                                                                                                                                                                                                                                                                                                                                                                                                                                                                                                               
	\end{tabular}
	\caption{Case study of the context generated using LLaMA-70b on LLaMA-7b in Few-Shot. Words in \textcolor{blue}{blue} indicate the correct answer, while those in \textcolor{red}{red} represent other potential answers.}
	\label{tbl:case_study_generative}
\end{table*}

\begin{table*}[p]
	\small
	\begin{tabular}{@{}p{\textwidth}@{}}
		\toprule                                                                                                                                                                                                                                                                                                                                                                                                                                                                                                                                                                                                                                                                                                                                                                                                                                                                                                                                                                                                                                                                           
		\textbf{Question:} what city of USA has a neighborhood called little havana?                                                                                                                                                                                                                                                                                                                                                                                                                                                                                                                                                                                                                                                                                                                                                                                                                                                                                                                                                                                                       \\
\bottomrule                                                                                                                                                                                                                                                                                                                                                                                                                                                                                                                                                                                                                                                                                                                                                                                                                                                                                                                                                                                                                                                                        
	\end{tabular}
	\caption{Case study of the hints generated using \method on LLaMA-7b in Few-Shot. Words in \textcolor{blue}{blue} indicate the correct answer.}
	\label{tbl:case_study_higenqa}
\end{table*}

\clearpage

\end{document}